%% file: iclr2018.tex
\documentclass{article} 

\usepackage{iclr2018_conference,times}

\usepackage{mathtools}
\usepackage{hyperref}
\usepackage{url}
\usepackage{paralist}
\usepackage[pdftex]{graphicx}
\usepackage{amsmath,amssymb}
\usepackage{verbatim}
\usepackage{mathabx}
\usepackage[export]{adjustbox}
\usepackage{multirow}
\usepackage{amsthm}

\newcommand{\ie}{i.e.,}
\newcommand{\eg}{e.g.,}

\newcommand{\argmax}{\operatornamewithlimits{argmax}}

\definecolor{myGreen}{rgb}{0.29, 0.60, 0.23}

\graphicspath{{images/}}
\DeclareGraphicsExtensions{.png,.jpg,.pdf}

\title{Interactive Grounded Language Acquisition and Generalization in
  a 2D World}

\author{Haonan Yu$^1$, Haichao Zhang$^1$ \& Wei Xu$^{1,2}$\\
$^1$Baidu Research, Sunnyvale USA\\
$^2$National Engineering Laboratory for Deep Learning Technology and
Applications, Beijing China\\
\texttt{\{haonanyu,zhanghaichao,wei.xu\}@baidu.com} \\
}

\iclrfinalcopy 

\begin{document}

\maketitle

\begin{abstract}
  We build a virtual agent for learning language in a 2D maze-like
  world.
  The agent sees images of the surrounding environment, listens to a
  virtual teacher, and takes actions to receive rewards.
  It interactively learns the teacher's language from scratch based
  on two language use cases: sentence-directed navigation and question
  answering.
  It learns simultaneously the visual representations of the
  world, the language, and the action control.
  By disentangling language grounding from other computational
  routines and sharing a concept detection function between language
  grounding and prediction, the agent reliably interpolates and
  extrapolates to interpret sentences that contain new word
  combinations or new words missing from training sentences.
  The new words are transferred from the answers of language
  prediction.
  Such a language ability is trained and evaluated on a
  population of over 1.6 million distinct sentences consisting of 119
  object words, 8 color words, 9 spatial-relation words, and 50
  grammatical words.
  The proposed model significantly outperforms five comparison
  methods for interpreting zero-shot sentences.
  In addition, we demonstrate human-interpretable intermediate
  outputs of the model in the appendix.
\end{abstract}

\section{Introduction}

Some empiricists argue that language may be learned based on its
usage~\citep{Tomasello2003}.
\citet{Skinner57} suggests that the successful use of a word reinforces
the understanding of its meaning as well as the probability of it
being used again in the future.
\citet{Bruner85} emphasizes the role of social interaction in helping
a child develop the language, and posits the importance of the
feedback and reinforcement from the parents during the learning
process.
This paper takes a positive view of the above behaviorism and tries to
explore some of the ideas by instantiating them in a 2D virtual
world where \emph{interactive} language acquisition happens.
This interactive setting contrasts with a common learning setting in
that language is learned from dynamic interactions with environments
instead of from static labeled data.

Language acquisition can go beyond mapping language as input patterns
to output labels for merely obtaining high rewards or accomplishing
tasks.
We take a step further to require the language to be
\emph{grounded}~\citep{Harnad90}.
Specifically, we consult the paradigm of procedural
semantics~\citep{Woods07} which posits that words, as abstract
procedures, should be able to pick out referents.
We will attempt to explicitly link words to environment concepts
instead of treating the whole model as a black box.
Such a capability also implies that, depending on the interactions with
the world, words would have particular meanings in a particular
context and some content words in the usual sense might not even have
meanings in our case.
As a result, the goal of this paper is to acquire ``in-context'' word
meanings regardless of their suitability in all scenarios.

On the other hand, it has been argued that a child's exposure to adult
language provides inadequate evidence for language
learning~\citep{Chomsky91}, but some induction mechanism should exist
to bridge this gap~\citep{Landauer97}.
This property is critical for any AI system to learn an infinite
number of sentences from a finite amount of training data.
This type of generalization problem is specially addressed in our
problem setting.
After training, we want the agent to generalize to interpret
\emph{zero-shot} sentences of two types:
\begin{compactenum}[1)]
\item \emph{interpolation}, new combinations of previously seen words
  for the same use case, or
\item \emph{extrapolation}, new words transferred from other use cases
  and models.
\end{compactenum}
In the following, we will call the first type ZS1 sentences and the
second type ZS2 sentences.
Note that so far the zero-shot problems, addressed by most recent
work~\citep{Hermann17,Chaplot18} of interactive language learning,
belong to the category of ZS1.
In contrast, a reliable interpretation of ZS2 sentences, which is
essentially a \emph{transfer learning} \citep{Pan2010} problem, will
be a major contribution of this work.

\begin{figure}
  \setlength{\tabcolsep}{0.5pt}
  \begin{center}
    \resizebox{\textwidth}{!}{
    \begin{tabular}{@{}c@{\hskip 4pt}cc|cc@{}}
      \multicolumn{3}{c|}{Training} & \multicolumn{2}{c}{Testing}\\
      &
      \includegraphics[width=0.24\textwidth]{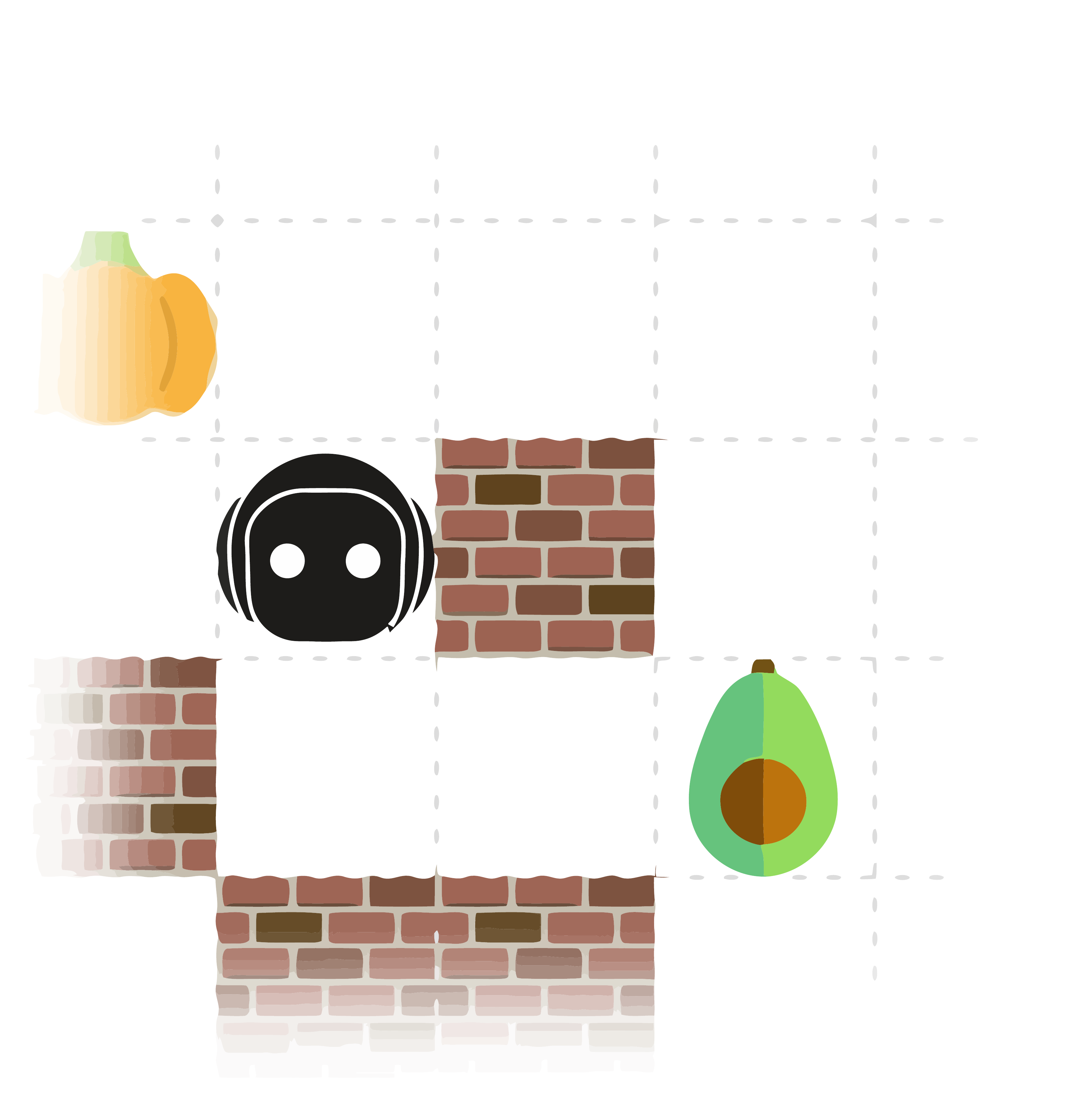} &
      \includegraphics[width=0.24\textwidth]{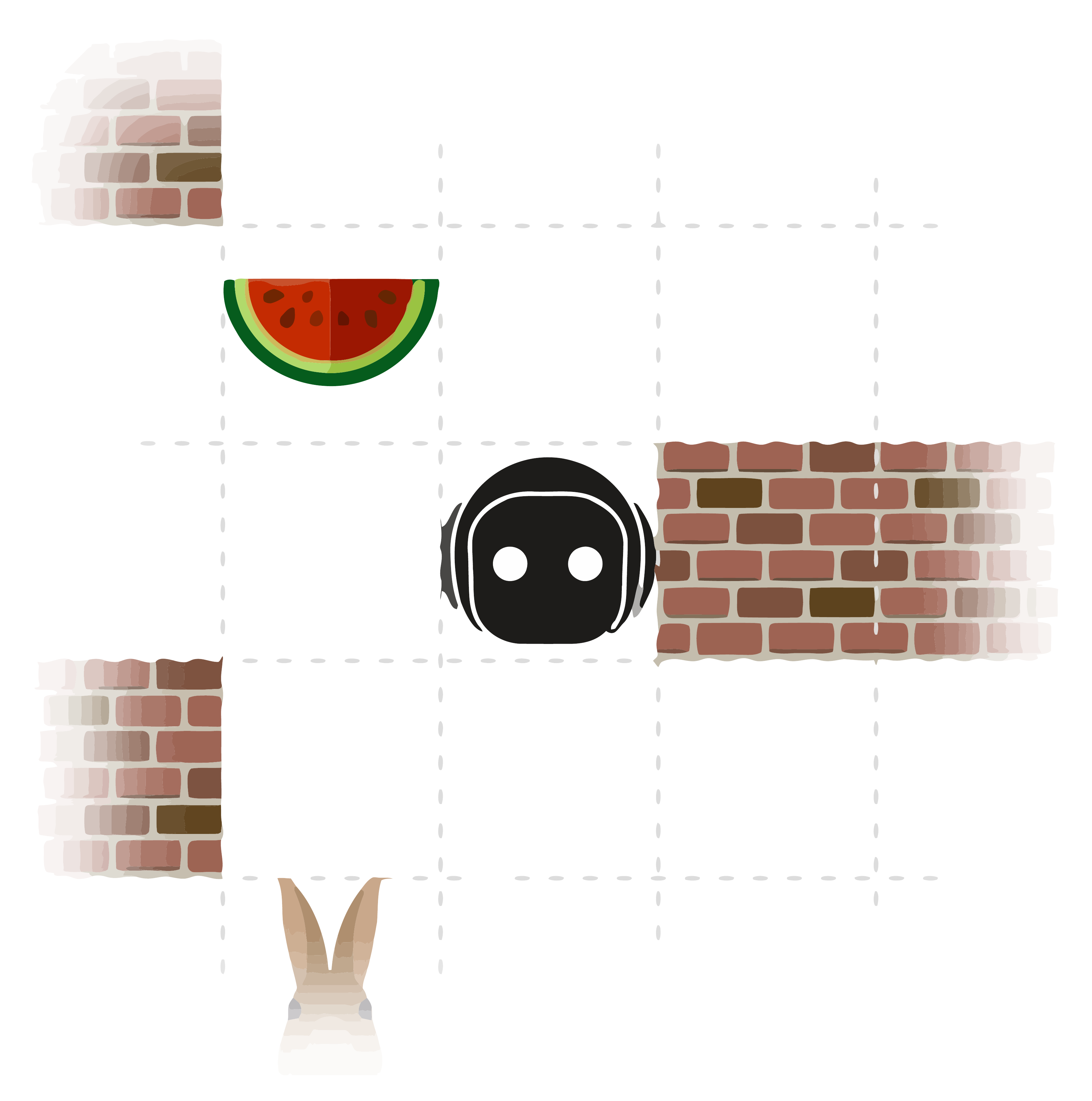} &
      \includegraphics[width=0.24\textwidth]{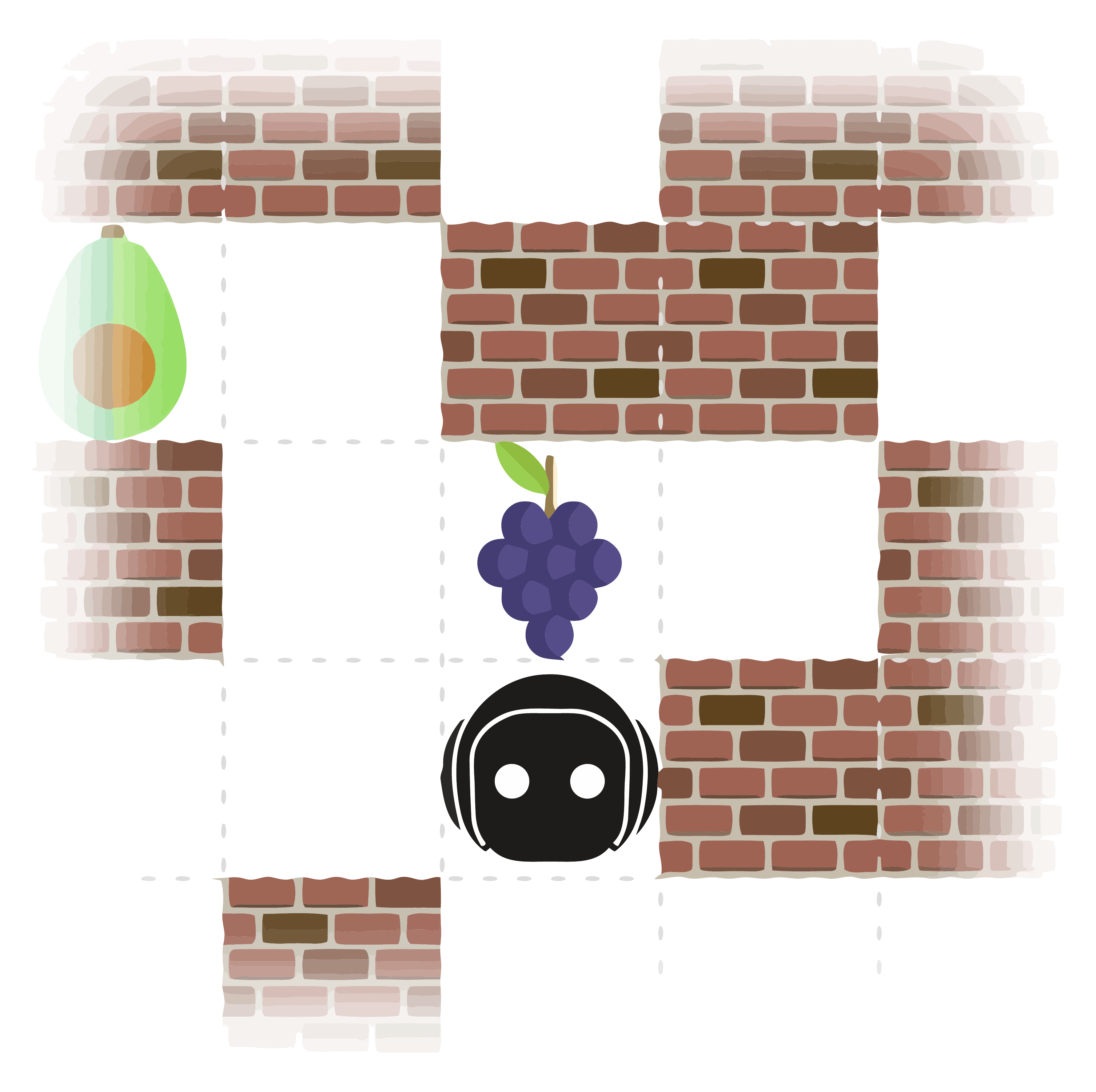} &
      \includegraphics[width=0.24\textwidth]{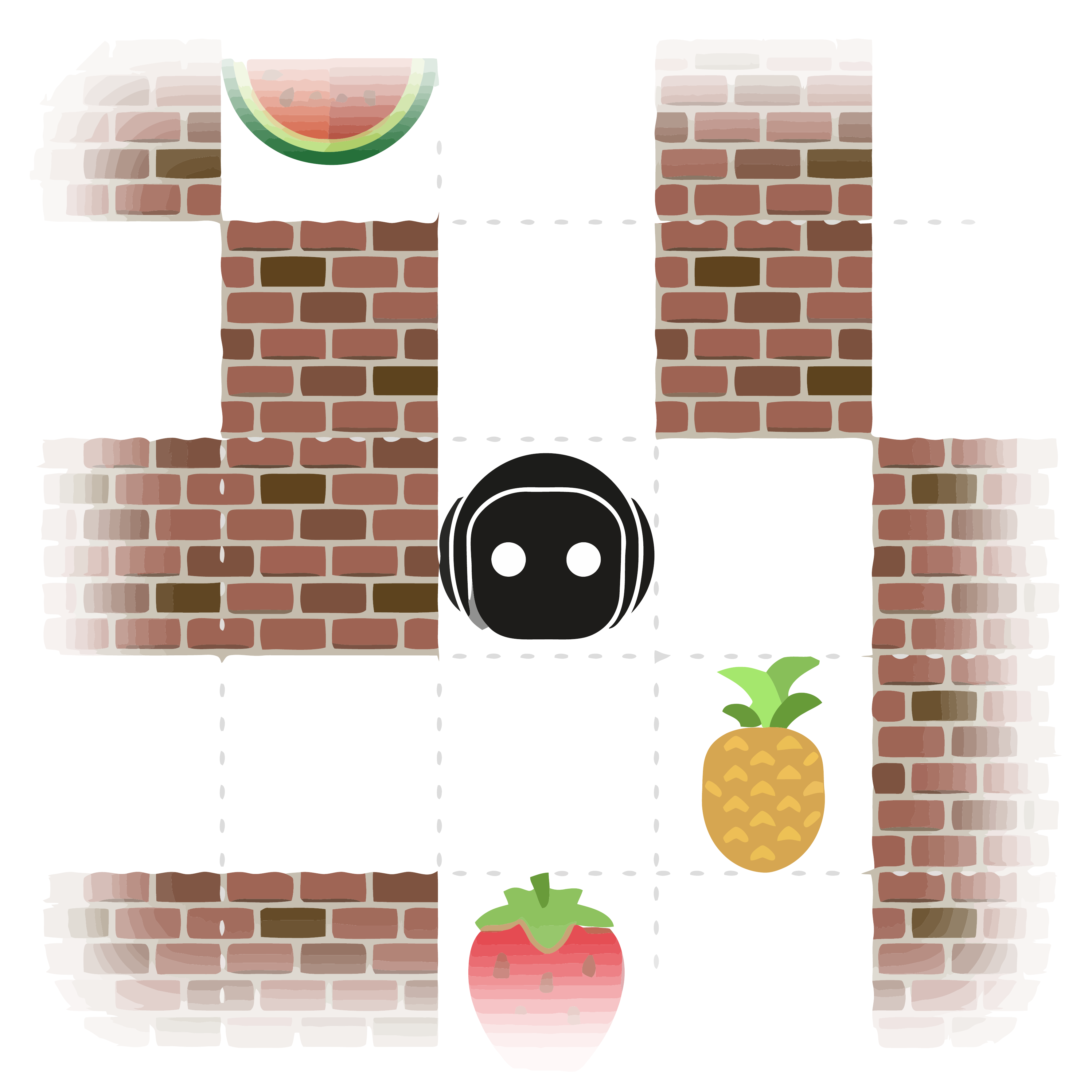}\\
      NAV &
      {\small``Move to north of {\color{blue}avocado}.''} &
      {\small``Go to {\color{blue}east} of rabbit.''} &
      {\small``Go to {\color{blue}east} of {\color{blue}avocado}.''} &
      {\small``Can you reach {\color{red}watermelon}?''}
      \\[2mm]
      QA &
      & {\small``What is in northwest?''}
      & {\small ``What is {\color{blue}east} of
          {\color{blue}avocado}?''}
      & {\small ``What is the color of {\color{red}watermelon}?''}
      \\
      &
      & {\small(\textit{Answer}: ``{\color{red}Watermelon}'')}
      & {\small(\textit{Answer}: ``Nothing'')}
      & {\small(\textit{Answer}: ``Red'')}
      \\
      & (a) & (b) & (c) & (d)\\
    \end{tabular}
    }
    \caption{An illustration of \textsc{xworld} and
      the two language use cases.
      (a) and (b): A mixed training of NAV and QA.
      (c): Testing ZS1 sentences contain a new combination of words
      (``east'' and ``avocado'') that never appear together in any
      training sentence.
      (d): Testing ZS2 sentences contain a new word
      (``watermelon'') that never appears in any training sentence
      but is learned from a training answer.
      This figure is only a conceptual illustration of language
      generalization; in practice it might take many training sessions
      before the agent can generalize.
      (Due to space limitations, the maps are only partially shown.)
    }
    \label{fig:task}
  \end{center}
\end{figure}

We created a 2D maze-like world called \textsc{xworld}
(Figure~\ref{fig:task}), as a testbed for interactive grounded
language acquisition and
generalization.\footnote{\url{https://github.com/PaddlePaddle/XWorld}}
In this world, a virtual agent has two language use cases:
navigation (NAV) and question answering (QA).
For NAV, the agent needs to navigate to correct places indicated by
language commands from a virtual teacher.
For QA, the agent must correctly generate single-word answers to the
teacher's questions.
NAV tests language comprehension while QA additionally tests language
prediction.
They happen simultaneously: When the agent is navigating, the teacher
might ask questions regarding its current interaction with the
environment.
Once the agent reaches the target or the time is up, the current
\emph{session} ends and a new one is randomly generated according to
our configuration (Appendix~\ref{app:xworld}).
The ZS2 sentences defined in our setting require word meanings to be
transferred from single-word answers to sentences, or more precisely,
\emph{from language prediction to grounding}.
This is achieved by establishing an explicit link between grounding
and prediction via a common concept detection function, which
constitutes the major novelty of our model.
With this transferring ability, the agent is able to comprehend a
question containing a new object learned from an answer, without
retraining the QA pipeline.
It is also able to navigate to a freshly taught object without
retraining the NAV pipeline.

It is worthwhile emphasizing that this seemingly ``simple'' world in
fact poses great challenges for language acquisition and
generalization, because:
\begin{compactenum}[$\circ$]
\item \emph{The state space is huge.}
Even for a $7\times7$ map with 15 wall blocks and 5 objects
selected from 119 distinct classes, there are already octillions
($10^{27}$) of possible different configurations, not to mention the
intra-class variance of object instances (see Figure~\ref{fig:icons}
in the appendix).
For two configurations that only differ in one block, their successful
navigation paths could be completely different.
This requires an accurate perception of the environment.
Moreover, the configuration constantly changes from session to
session, and from training to testing.
In particular, the target changes across sessions in both location and
appearance.
\item \emph{The goal space implied by the language for navigation is
  huge.}
For a vocabulary containing only 185 words, the total number of
distinct commands that can be said by the teacher conforming to our
defined grammar is already over half a million.
Two commands that differ by only one word could imply completely
different goals.
This requires an accurate grounding of language.
\item \emph{The environment demands a strong language generalization
  ability from the agent.}
The agent has to learn to interpret zero-shot sentences that might be
as long as 13 words.
It has to ``plug'' the meaning of a new word or word combination into
a familiar sentential context while trying to still make sense of the
unfamiliar whole.
The recent work~\citep{Hermann17,Chaplot18} addresses ZS1 (for short
sentences with several words) but not ZS2 sentences, which is a key
difference between our learning problem and theirs.
\end{compactenum}

We describe an end-to-end model for the agent to interactively acquire
language from scratch and generalize to unfamiliar sentences.
Here ``scratch'' means that the model does not hold any assumption of
the language semantics or syntax.
Each sentence is simply a sequence of tokens with each token being
equally meaningless in the beginning of learning.
This is unlike some early pioneering systems
(\eg\ SHRDLU~\citep{Winograd72} and
\textsc{Abigail}~\citep{Siskind1994}) that hard-coded the syntax or
semantics to link language to a simulated world--an approach that
presents scalability issues.
There are two aspects of the interaction: one is with the teacher
(\ie\ language and rewards) and the other is with the environment
(\eg\ stepping on objects or hitting walls).
The model takes as input RGB images, sentences, and rewards.
It learns simultaneously the visual representations of the world, the
language, and the action control.
We evaluate our model on randomly generated \textsc{xworld} maps with
random agent positions, on a population of over 1.6 million distinct
sentences consisting of 119 object words, 8 color words, 9
spatial-relation words, and~50 grammatical words.
Detailed analysis (Appendix~\ref{sec:visualize}) of the trained model
shows that the language is grounded in such a way that the words are
capable to pick out referents in the environment.
We specially test the generalization ability of the agent for
handling zero-shot sentences.
The average NAV success rates are 84.3\% for ZS1 and 85.2\% for ZS2
when the zero-shot portion is half, comparable to the rate of 90.5\%
in a normal language setting.
The average QA accuracies are 97.8\% for ZS1 and 97.7\% for ZS2 when
the zero-shot portion is half, almost as good as the accuracy of
99.7\% in a normal language setting.

\section{Model}
Our model incorporates two objectives.
The first is to maximize the cumulative reward of NAV and the second
is to minimize the classification cost of QA.
For the former, we follow the standard reinforcement learning (RL)
paradigm: the agent learns the action at every step from reward
signals.
It employs the actor-critic (AC) algorithm~\citep{Sutton1998} to learn
the control policy (Appendix~\ref{app:ac}).
For the latter, we adopt the standard supervised setting of Visual
QA~\citep{Antol2015}: the groundtruth answers are provided by the
teacher during training.
The training cost is formulated as the multiclass cross entropy.

\subsection{Motivation}
The model takes two streams of inputs: images and sentences.
The key is how to model the language grounding problem.
That is, the agent must link (either implicitly or explicitly)
language concepts to environment entities to correctly take an action
by understanding the instruction in the current visual context.
A straightforward idea would be to encode the sentence~$s$ with an RNN
and encode the perceived image~$e$ with a CNN, after which the two
encoded representations are mixed together.
Specifically, let the multimodal module be $\mathbf{M}$, the action
module be $\mathbf{A}$, and the prediction module be $\mathbf{P}$,
this idea can be formulated as:
\begin{equation}
  \label{eq:joint-embedding}
  \begin{array}{ll}
    \text{NAV:} & \mathbf{A}\big{(}\mathbf{M}(\text{RNN}(s), \text{CNN}(e))\big{)}\\
    \text{QA:} & \mathbf{P}\big{(}\mathbf{M}(\text{RNN}(s), \text{CNN}(e))\big{)}.\\
  \end{array}
\end{equation}
\citet{Hermann17,Misra17,Chaplot18} all employ the above paradigm.
In their implementations, $\mathbf{M}$ is either vector concatenation
or element-wise product.
For any particular word in the sentence, fusion with the image could
happen anywhere starting from $\mathbf{M}$ all the way to the end,
right before a label is output.
This is due to the fact that the RNN folds the string of words into a
compact embedding which then goes through the subsequent blackbox
computations.
Therefore, language grounding and other computational routines are
entangled.
Because of this, we say that this paradigm has an \emph{implicit}
language grounding strategy.
Such a strategy poses a great challenge for processing a ZS2 sentence
because it is almost impossible to predict how a new word learned from
language prediction would perform in the complex entanglement
involved.
Thus a careful inspection of the grounding process is needed.

\subsection{Approach}
\begin{figure}[t]
  \begin{center}
    \resizebox{0.95\textwidth}{!}{
      \includegraphics[width=\textwidth]{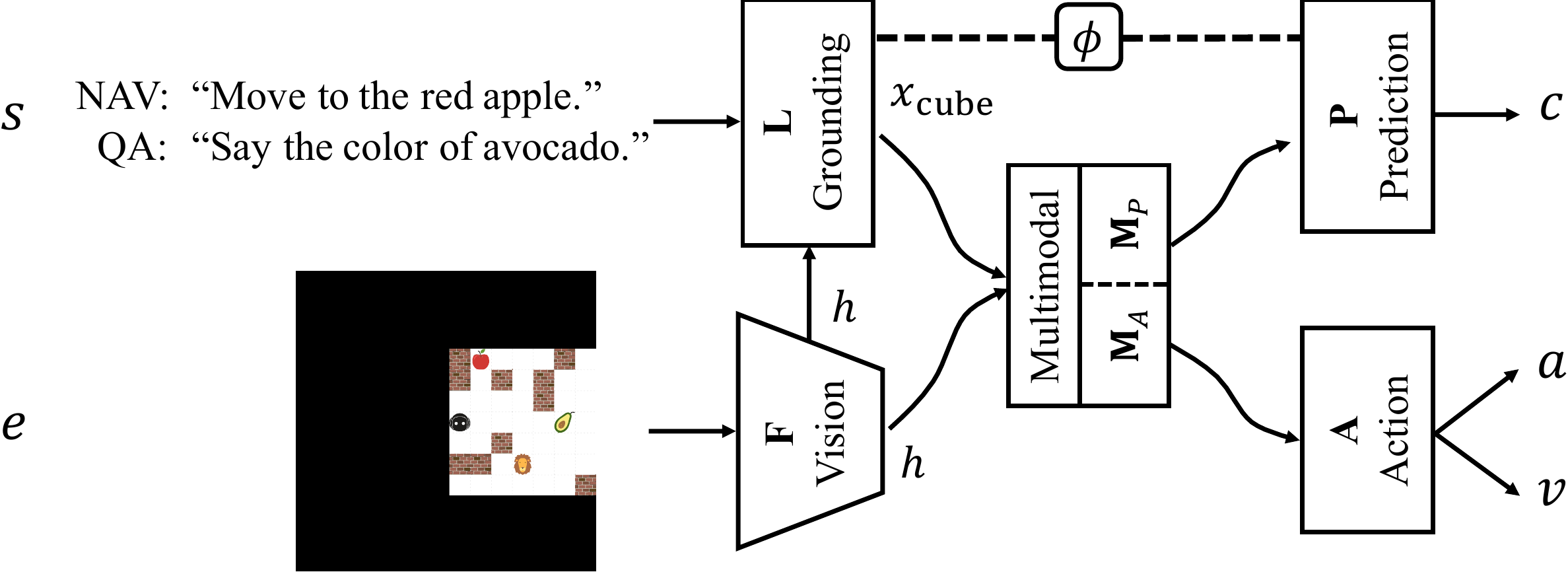}
    }
    \caption{An overview of the model.
      We process $e$ by always placing the agent at the center via zero
      padding.
      This helps the agent learn navigation actions by reducing the
      variety of target representations.
      $c$, $a$, and~$v$ are the predicted answer, the navigation
      action, and the critic value for policy gradient, respectively.
      $\phi$ denotes the concept detection function shared by
      language grounding and prediction.
      $\mathbf{M}_A$ generates a compact representation from $x_{\text{loc}}$
      and $h$ for navigation (Appendix~\ref{app:details}).
    }
    \label{fig:overview}
  \end{center}
\end{figure}

The main idea behind our approach is to \emph{disentangle} language
grounding from other computations in the model.
This disentanglement makes it possible for us to explicitly define
language grounding around a core function that is also used by
language prediction.
Specifically, both grounding and prediction are cast as concept
detection problems, where each word (embedding) is treated as a
detector.
This opens up the possibility of transferring word meanings from the
latter to the former.
The overall architecture of our model is shown in
Figure~\ref{fig:overview}.

\subsubsection{Explicit grounding}
\label{sec:grounding}
We begin with our definition of ``grounding.''
We define a sentence as generally a string of words of any length.
A single word is a special case of a sentence.
Given a sentence~$s$ and an image representation~$h=\text{CNN}(e)$, we
say that $s$ is \emph{grounded} in $h$ as~$x$ if
\begin{compactenum}[I)]
\item $h$ consists of $M$ entities where an entity is a subset of
  visual features, and
\item $x\in\{0,1\}^M$ with each entry~$x[m]$ representing a binary selection
  of the $m$th entity of~$h$.
  Thus $x$ is a combinatorial selection over $h$.
\end{compactenum}
Furthermore, $x$ is \emph{explicit} if

\ \ III) it is formed by the grounding results of (some) individual
words of $s$ (\ie\ compositionality).

We say that a framework has an explicit grounding strategy if

\ \ IV) \emph{all} language-vision fusions in the framework are
explicit groundings.

For our problem, we propose a new framework with an explicit grounding strategy:
\begin{equation}
  \label{eq:explicit-grounding}
  \begin{array}{ll}
    \text{NAV:} & \mathbf{A}\big{(}\mathbf{M}_A(x, \text{CNN}(e))\big{)}\\
    \text{QA:} & \mathbf{P}\big{(}\mathbf{M}_P(x, \text{CNN}(e))\big{)},\\
  \end{array}
\end{equation}
where the sole language-vision fusion $x$ in the framework is an
explicit grounding.
Notice in the above how the grounding process, as a ``bottleneck,''
allows only $x$ but not other linguistic information to flow to the
downstream of the network.
That is, $\mathbf{M}_A$, $\mathbf{M}_P$, $\mathbf{A}$, and
$\mathbf{P}$ \emph{all} rely on grounded results but \emph{not} on
other sentence representations.
By doing so, we expect $x$ to summarize all the necessary linguistic
information for performing the tasks.

The benefits of this framework are two-fold.
First, the explicit grounding strategy provides a \emph{conceptual
  abstraction}~\citep{Garnelo2016} that maps high-dimensional
linguistic input to a lower-dimensional conceptual state space and
abstracts away irrelevant input signals.
This improves the generalization for similar linguistic inputs.
Given $e$, all that matters for NAV and QA is $x$.
This guarantees that the agent will perform exactly in the same way
on the same image $e$ even given different sentences as long as their
grounding results $x$ are the same.
It disentangles language grounding from subsequent computations such
as obstacle detection, path planning, action making, and feature
classification, which all should be inherently language-independent
routines.
Second, because $x$ is explicit, the roles played by the individual
words of $s$ in the grounding are interpretable.
This is in contrast to Eq.~\ref{eq:joint-embedding} where the roles of
individual words are unclear.
The interpretability provides a possibility of establishing a link
between language grounding and prediction, which we will perform in
the remainder of this section.

\subsubsection{Instantiation of explicit grounding}
Let $h\in\mathbb{R}^{N\times D}$ be a spatially flattened feature cube
(originally in 3D, now the 2D spatial domain collapsed into 1D for
notational simplicity), where~$D$ is the number of channels and $N$ is
the number of locations in the spatial domain.
We adopt three definitions for an entity:
\begin{compactenum}[1)]
\item a feature vector at a particular image location,
\item a particular feature map along the channel dimension, and
\item a scalar feature at the intersection of a feature vector and a
  feature map.
\end{compactenum}
Their grounding results are denoted as $x_{\text{loc}}(s,h)\in\{0,1\}^N$,
$x_{\text{feat}}(s,h)\in\{0,1\}^D$, and $x_{\text{cube}}(s,h)\in\{0,1\}^{N\times
  D}$, respectively.
In the rest of the paper, we remove $s$ and $h$ from $x_{\text{loc}}$,
$x_{\text{feat}}$, and $x_{\text{cube}}$ for notational simplicity while always
assuming a dependency on them.
We assume that $x_{\text{cube}}$ is a low-rank matrix that can be decomposed
into the two:
\begin{eqnarray*}
  \begin{split}
    x_{\text{cube}} &=x_{\text{loc}}\cdot{x_{\text{feat}}}^{\intercal}.\\
  \end{split}
\end{eqnarray*}

To make the model fully differentiable, in the following we relax the
definition of grounding so that $x_{\text{loc}}\in[0,1]^N$,
$x_{\text{feat}}\in[0,1]^D$, and~$x_{\text{cube}}\in[0,1]^{N\times D}$.
The attention map~$x_{\text{loc}}$ is responsible for image spatial
attention.
The channel mask~$x_{\text{feat}}$ is responsible for selecting image feature
maps, and is assumed to be independent of the specific~$h$, namely,
$x_{\text{feat}}(s,h)=x_{\text{feat}}(s)$.
Intuitively, $h$ can be modulated by $x_{\text{feat}}$ before being sent to
downstream processings.
A recent paper by \citet{Vries2017} proposes an even earlier modulation of
the visual processing by directly conditioning some of the parameters
of a CNN on the linguistic input.

Finally, we emphasize that our explicit grounding, even though
instantiated as a soft attention mechanism, is different from the
existing visual attention models.
Some attention models such as \citet{Xu15, Vries2017} violate
definitions III and IV.
Some work \citep{Andreas2016a,Andreas2016b,Lu2016} violates definition
IV in a way that language is fused with vision by a multilayer
perceptron (MLP) after image attention.
\citet{Anderson2017} proposes a pipeline similar to ours but violates
definition III in which the image spatial attention is computed from a
compact question embedding output by an RNN.

\subsubsection{Concept detection}
With language grounding disentangled, now we relate it to language
prediction.
This relation is a common \emph{concept detection} function.
We assume that every word in a vocabulary, as a concept, is detectable
against entities of type (1) as defined in Section~\ref{sec:grounding}.
For a meaningful detection of spatial-relation words that are
irrelevant to image content, we incorporate parametric feature maps
into~$h$ to learn spatial features.
Assume a precomputed $x_{\text{feat}}$, the concept detection operates by sliding
over the spatial domain of the feature cube~$h$, which can be written
as a function~$\phi$:
\[\phi : h, x_{\text{feat}}, u \mapsto \chi,\]
where~$\chi\in\mathbb{R}^N$ is a detection score map and $u$ is a word
embedding vector.
This function scores the embedding~$u$ against each feature vector
of~$h$, modulated by~$x_{\text{feat}}$ that selects which feature maps to use for
the scoring.
Intuitively, each score on $\chi$ indicates the detection response of
the feature vector in that location.
A higher score represents a higher detection response.

While there are many potential forms for~$\phi$, we implement it as
\begin{equation}
  \label{eq:phi}
  \phi(h,x_{\text{feat}},u)=h\cdot(x_{\text{feat}}\circ u),
\end{equation}
where~$\circ$ is the element-wise product.
To do so, we have word embedding~$u\in\mathbb{R}^D$ where $D$ is equal
to the number of channels of~$h$.

\subsubsection{Prediction by concept detection}
For prediction, we want to output a word given a question $s$ and
an image $e$.
Suppose that $x_{\text{loc}}$ and $x_{\text{feat}}$ are the grounding results of $s$.
Based on the detection function $\phi$, $\mathbf{M}_P$ outputs a score
vector~$m\in\mathbb{R}^K$ over the entire lexicon, where each entry of
the vector is:
\begin{eqnarray}
  \label{eq:m}
    m[k]=x_{\text{loc}}^{\intercal}\phi(h,x_{\text{feat}},u_k)=x_{\text{loc}}^{\intercal}\chi_k,
\end{eqnarray}
where~$u_k$ is the $k$th entry of the word embedding table.
The above suggests that~$m[k]$ is the result of weighting the
scores on the map $\chi_k$ by $x_{\text{loc}}$.
It represents the correctness of the $k$th lexical entry as the
answer to the question~$s$.
To predict an answer
\[\mathbf{P}(m)=\argmax_k\big(\text{softmax}(m)\big).\]
Note that the role of~$x_{\text{feat}}$ in the prediction is to select which
feature maps are relevant to the question~$s$.
Otherwise it would be confusing for the agent about what to predict
(\eg\ whether to predict a color or an object name).
By using $x_{\text{feat}}$, we expect that different feature maps encode
different image attributes (see an example in the caption of
Figure~\ref{fig:attention}).
More analysis of $x_{\text{feat}}$ is performed in Appendix~\ref{sec:visualize}.

\begin{figure}[t]
  \begin{center}
    \begin{tabular}{c}
      \resizebox{0.5\textwidth}{!}{
        \includegraphics[width=\textwidth]{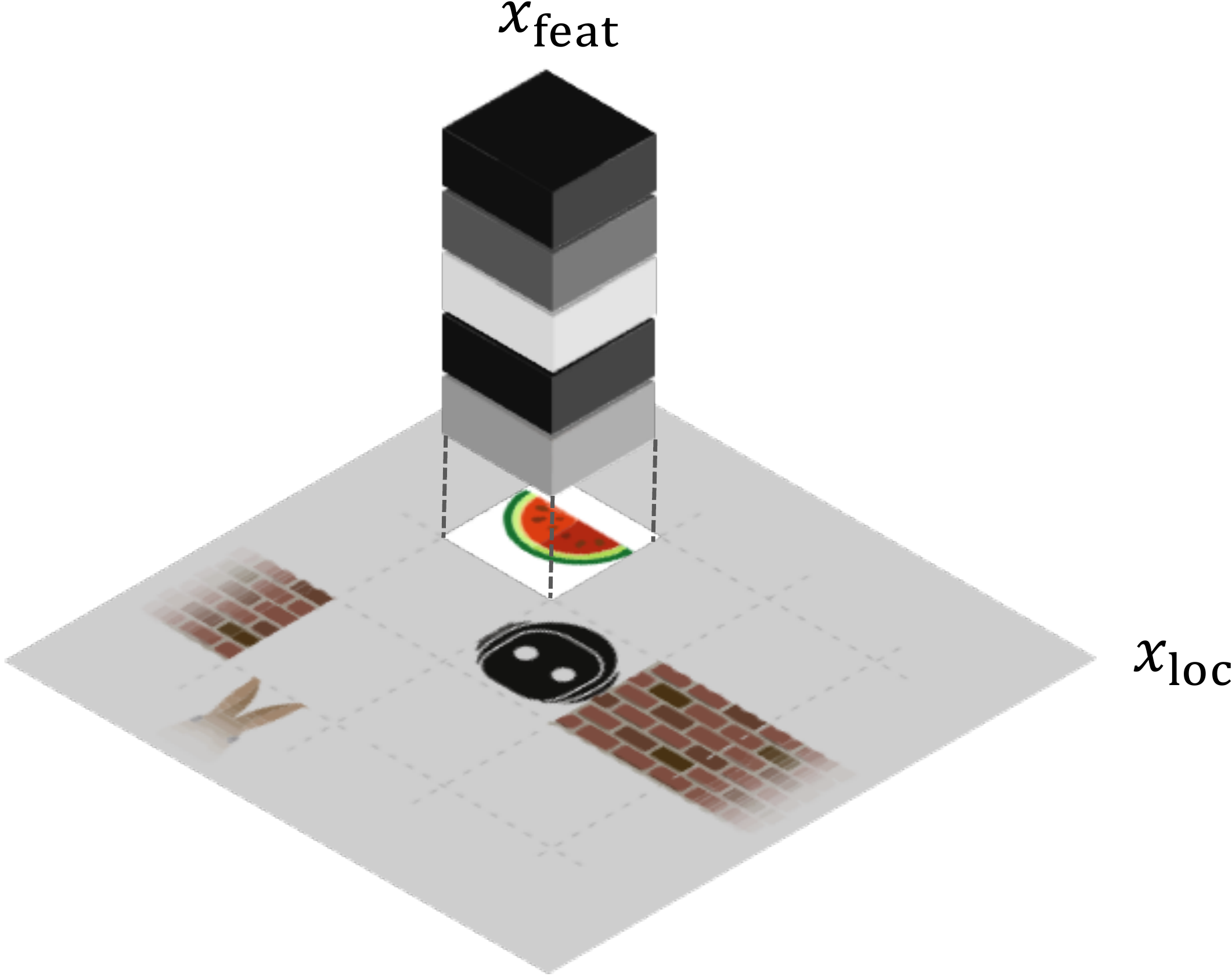}
      }\\
      ``What is the color of the object in the northeast?''\\
    \end{tabular}
    \caption{An illustration of the attention cube
      $x_{\text{cube}}=x_{\text{loc}}\cdot{x_{\text{feat}}}^{\intercal}$, where~$x_{\text{loc}}$ attends
      to image regions and~$x_{\text{feat}}$ selects feature maps.
      In this example, $x_{\text{loc}}$ is computed from ``northeast.''
      In order for the agent to correctly answer ``red'' (color)
      instead of ``watermelon'' (object name), $x_{\text{feat}}$ has to be
      computed from the sentence pattern ``What ... color ...?''
    }
    \label{fig:attention}
  \end{center}
\end{figure}

\subsubsection{Grounding by concept detection}
To compute~$x_{\text{cube}}$, we compute~$x_{\text{loc}}$ and~$x_{\text{feat}}$ separately.

We want~$x_{\text{loc}}$ to be built on the detection function~$\phi$.
One can expect to compute a series of score maps~$\chi$ of individual
words and merge them into~$x_{\text{loc}}$.
Suppose that $s$ consists of $L$ words $\{w_l\}$ with $w_l=u_k$ being
some word $k$ in the dictionary.
Let $\tau(s)$ be a sequence of indices $\{l_i\}$ where $0 \le l_i <
L$.
This sequence function~$\tau$ decides which words of the sentence are
selected and organized in what order.
We define $x_{\text{loc}}$ as
\begin{eqnarray}
  \label{eq:x_loc}
  \begin{split}
    x_{\text{loc}} &=
    \displaystyle\Upsilon\big{(}\phi(h,\mathbf{1},w_{l_1}),\ldots,\phi(h,\mathbf{1},w_{l_i}),\ldots,\phi(h,\mathbf{1},w_{l_I})\big{)}\\
    &=\displaystyle\Upsilon(\chi_{l_1},\ldots,\chi_{l_i},\ldots,\chi_{l_I}),
  \end{split}
\end{eqnarray}
where~$\mathbf{1}\in\{0,1\}^D$ is a vector of ones, meaning that it
selects all the feature maps for detecting~$w_{l_i}$.
$\Upsilon$ is an aggregation function that combines the sequence of
score maps~$\chi_{l_i}$ of individual words.
As such, $\phi$ makes it possible to transfer new words from
Eq.~\ref{eq:m} to Eq.~\ref{eq:x_loc} during test time.

If we were provided with an oracle that is able to output a parsing
tree for any sentence, we could set $\tau$ and~$\Upsilon$ according to
the tree semantics.
Neural module networks (NMNs)~\citep{Andreas2016a,Andreas2016b,Hu2017}
rely on such a tree for language grounding.
They generate a network of modules where each module corresponds to a
tree node.
However, labeled trees are needed for training.
Below we propose to learn $\tau$ and~$\Upsilon$ based on word
attention~\citep{Bahdanau2014} to bypass the need for labeled
structured data.

\begin{figure}[t]
  \begin{center}
    \resizebox{0.75\textwidth}{!}{
      \includegraphics[width=\textwidth]{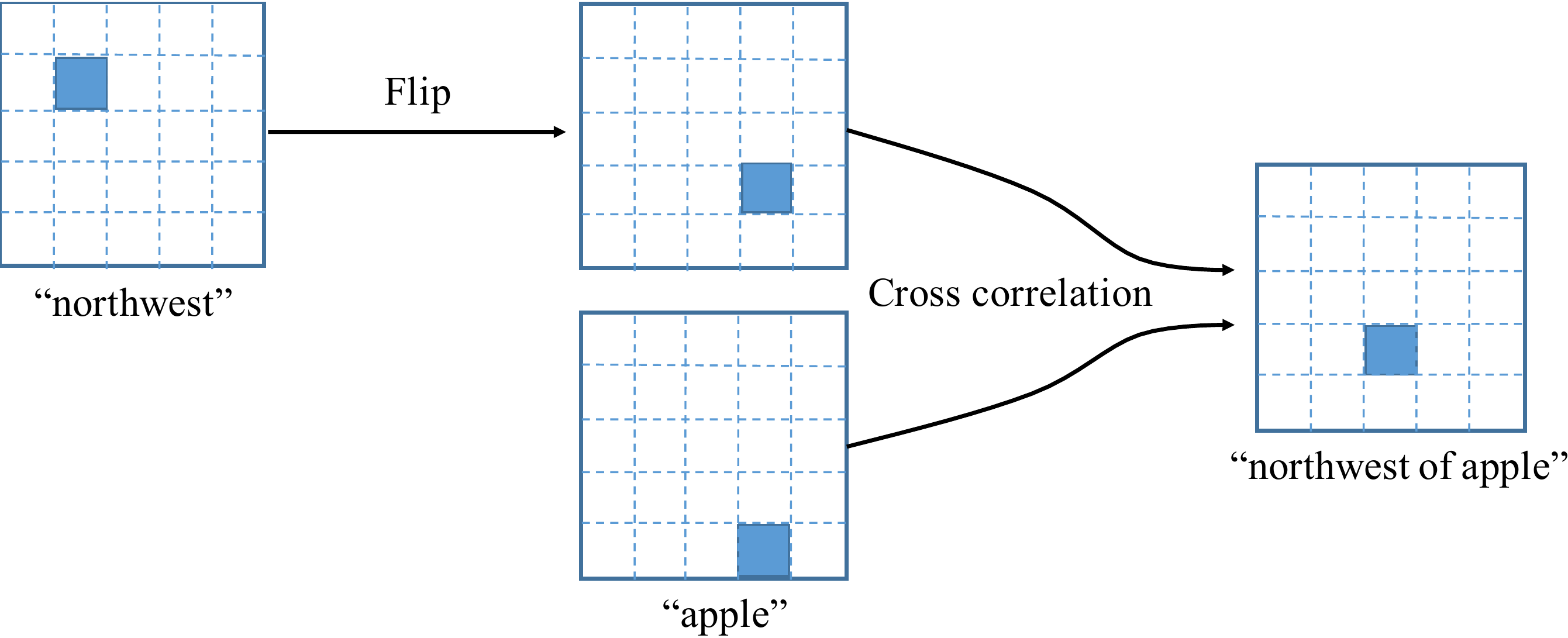}
    }
    \caption{A symbolic example of the 2D convolution for transforming
      attention maps.
      A 2D convolution can be decomposed into two steps: flipping and
      cross correlation.
      The attention map of ``northwest'' is treated as an offset
      filter to translate that of ``apple.''
      Note that in practice, the attention is continuous and noisy,
      and the interpreter has to learn to find out the words (if any)
      to perform this convolution.
    }
    \label{fig:convolution}
  \end{center}
\end{figure}

We start by feeding a sentence~$s=\{w_l\}$ of length~$L$ to a
bidirectional RNN~\citep{Schuster1997}.
It outputs a compact sentence embedding~$s_{\text{emb}}$ and a
sequence of $L$ word context vectors~$\overline{w}_l$.
Each $\overline{w}_l$ summarizes the sentential pattern around that word.
We then employ a meta controller called \emph{interpreter} in an
iterative manner.
For the $i$th interpretation step, the interpreter computes the word
attention as:
\begin{equation}
  \label{eq:attention}
  \begin{array}{ll}
    \tau^*
    &
    \left\{
    \begin{array}{rl@{\hskip 3pt}l}
      \text{Word attention:} &
      o^i_l\ \ \propto & \exp\left[S_{\text{cos}}(p^{i-1}, \overline{w}_l)\right]\\
      \text{Attended context:} &
      \overline{w}^i= & \displaystyle\sum_lo^i_l\overline{w}_l\\
      \text{Attended word:} &
      s^i= & \displaystyle\sum_lo^i_lw_l\\
      \text{Interpreter state:} & p^i= &
      \text{GRU}(p^{i-1},\overline{w}^i)\\
    \end{array}
    \right.
    \\
  \end{array}
\end{equation}
where~$S_{\text{cos}}$ is cosine similarity and GRU is the gated recurrent
unit~\citep{Cho2014}.
Here we use $\tau^*$ to represent an approximation of~$\tau$ via soft
word attention.
We set~$p^0$ to the compact sentence embedding~$s_{\text{emb}}$.
After this, the attended word~$s^i$ is fed to the detection
function~$\phi$.
The interpreter aggregates the score map of $s^i$ by:
\begin{equation}
  \label{eq:aggregation}
  \begin{array}{ll}
    \Upsilon
    &
    \left\{
    \begin{array}{rl@{\hskip 3pt}l}
      \text{Detection:} & y'= &
      \text{softmax}\big{(}\phi(h,\mathbf{1},s^i)\big{)}\\
      \text{Map transform:} & x_{\text{loc}}^i= &
      y' * y^{i-1}\\
      \text{Map update gate:} & \rho^i= &
      \sigma(Wp^i + b)\\
      \text{Map update:} & y^i= &
      \rho^i x_{\text{loc}}^i + (1-\rho^i)y^{i-1}\\
    \end{array}
    \right.
    \\
  \end{array}
\end{equation}
where~$*$ denotes a 2D convolution, $\sigma$ is sigmoid, and~$\rho^i$ is
a scalar.
$W$ and~$b$ are parameters to be learned.
Finally, the interpreter outputs~$x_{\text{loc}}^I$ as $x_{\text{loc}}$, where~$I$
is the predefined maximum step.

Note that in the above we formulate the map transform as a 2D
convolution.
This operation enables the agent to reason about spatial relations.
Recall that each attention map~$x_{\text{loc}}$ is egocentric.
When the agent needs to attend to a region specified by a spatial
relation referring to an object, it can translate the object attention
with the attention map of the spatial-relation word which serves as a
2D convolutional offset filter (Figure~\ref{fig:convolution}).
For this reason, we set~$y^0$ as a one-hot map where the map center
is one, to represent the identity translation.
A similar mechanism of spatial reasoning via convolution was explored
by \citet{Kitaev2017} for a voxel-grid 3D representation.

By assumption, the channel mask~$x_{\text{feat}}$ is meant to be determined
solely from~$s$; namely, which features to use should only depend on
the sentence itself, not on the value of the feature cube~$h$.
Thus it is computed as
\begin{equation}
  \label{eq:x_feat}
  x_{\text{feat}}=\text{MLP}\big{(}\text{RNN}(s)\big{)},
\end{equation}
where the RNN returns an average state of processing~$s$, followed by
an MLP with the sigmoid activation.\footnote{
  Note that here we drop the explicitness requirement
  (definition III) for Eq.~\ref{eq:x_feat}.
  This choice of implementation simplicity is purely based on our
  current problem that requires little language compositionality when
  computing $x_{\text{feat}}$ (unlike $x_{\text{loc}}$).
  One could imagine an alternative grounding that is explicit where a
  single-step word attention extracts words from $s$ to compute
  $x_{\text{feat}}$.
}

\section{Related work}
Our \textsc{xworld} is similar to the 2D block world MaseBase
\citep{Sukhbaatar2016}.
However, we emphasize the problem of grounded language acquisition and
generalization, while they do not.
There have been several 3D simulated worlds such as
ViZDoom~\citep{Kempka2016}, DeepMind Lab~\citep{Beattie2016}, and
Malmo~\citep{Johnson2016}.
Still, these other settings intended for visual perception and
control, with less or no language.

Our problem setting draws inspirations from the AI roadmap delineated
by \citet{Mikolov2015}.
Like theirs, we have a teacher in the environment that assigns tasks
and rewards to the agent, potentially with a curriculum.
Unlike their proposal of using only the linguistic channel, we have
multiple perceptual modalities, the fusion of which is believed to be
the basis of meaning~\citep{Kiela2016}.

Contemporary to our work, several end-to-end systems also address
language grounding problems in a simulated world with deep RL.
\citet{Misra17} maps instructions and visual observations to actions
of manipulating blocks on a 2D plane.
\citet{Hermann17,Chaplot18} learn to navigate in 3D under
instructions, and both evaluate ZS1 generalization.
Despite falling short of the vision challenges found in these other
worlds, we have a much larger space of zero-shot sentences and
additionally require ZS2 generalization, which is in fact a
transfer learning \citep{Pan2010} problem.

Other recent work~\citep{Andreas17,Oh17} on zero-shot multitask
learning treats language tokens as (parsed) labels for identifying
skills.
In these papers, the zero-shot settings are not intended for language
understanding.

The problem of grounding language in perception can perhaps be traced
back to the early work of \citet{Siskind1994,Siskind1999}, although no
statistical learning was adopted at that time.
Our language learning problem is related to some recent work on
learning to ground language in images and
videos~\citep{Yu2013,Gao2016,Rohrbach2016}.
The navigation task is relevant to robotics navigation under language
commands~\citep{Chen2011,Tellex2011,Barrett2017}.
The question-answering task is relevant to image question answering
(VQA)~\citep{Antol2015,Gao2015,Ren2015,Lu2016,Yang2016,Anderson2017,Vries2017}.
The interactive setting of learning to accomplish tasks is similar to
that of learning to play video games from pixels~\citep{Mnih2015}.

\section{Experiments}
We design a variety of experiments to evaluate the agent's language
acquisition and generalization ability.
Our model is first compared with several methods to demonstrate the
challenges in \textsc{xworld}.
We then evaluate the agent's language generalization ability in two
different zero-shot conditions.
Finally, we conclude with preliminary thoughts on how to scale our
model to a 3D world.

\begin{figure}[!t]
  \begin{center}
    \includegraphics[width=0.95\textwidth]{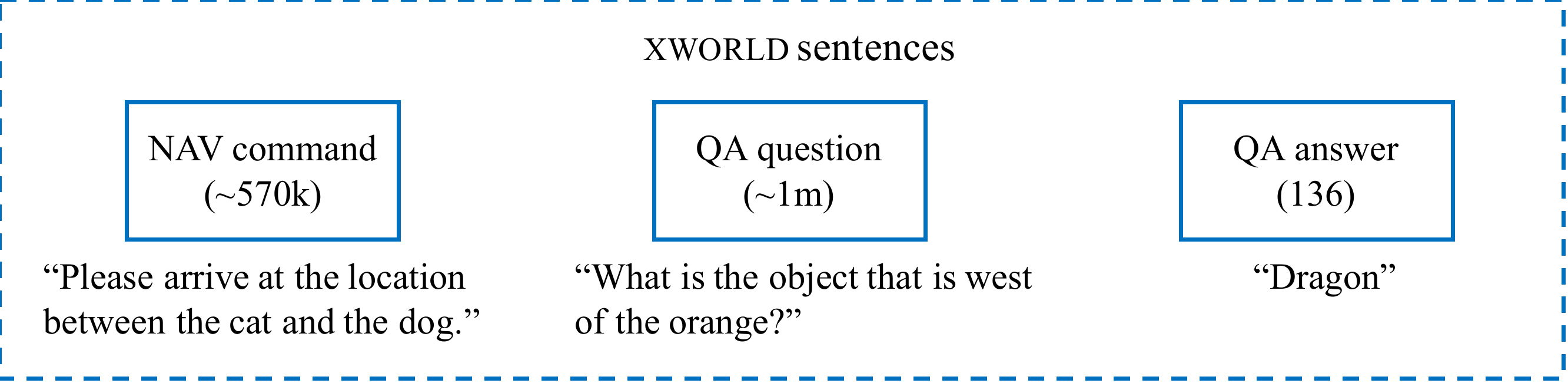}
  \end{center}
  \caption{The three types of language data and their statistics.
  }
  \label{fig:language}
\end{figure}

\subsection{General setup}
For all the experiments, both the sentences and the environments change
from session to session, and from training to testing.
The sentences are drawn conforming to the teacher's grammar.
There are three types of language data: NAV command, QA question, and
QA answer, which are illustrated in Figure~\ref{fig:language}.
In total, there are $\sim$570k NAV commands, $\sim$1m QA questions,
and 136 QA answers (all the content words plus ``nothing'' and minus
``between'').
The environment configurations are randomly generated from octillions
of possibilities of a $7\times7$ map, conforming to some high-level
specifications such as the numbers of objects and wall blocks.
For NAV, our model is evaluated on four types of navigation commands:

\begin{table*}[h!]
  \centering
  \begin{tabular}{ll}
    \texttt{nav\_obj}: & Navigate to an object.\\
    \texttt{nav\_col\_obj}: & Navigate to an object with a specific color.\\
    \texttt{nav\_nr\_obj}: & Navigate to a location near an object.\\
    \texttt{nav\_bw\_obj}: & Navigate to a location between two objects.\\
  \end{tabular}
\end{table*}
For QA, our model is evaluated on twelve types of questions
(\texttt{rec\_*} in Table~\ref{tab:sentence-types}).
We refer the reader to Appendix~\ref{app:xworld} for a detailed
description of the experiment settings.

\subsection{Comparison methods}
Four comparison methods and one ablation method are described below:

\noindent \textit{ContextualAttention} \textbf{[CA]} A variant of our
model that replaces the interpreter with a contextual attention model.
This attention model uses a gated RNN to convert a sentence to a filter
which is then convolved with the feature cube~$h$ to obtain the
attention map~$x_{\text{loc}}$.
The filter covers the $3\times 3$ neighborhood of each feature vector
in the spatial domain.
The rest of the model is unchanged.

\noindent\textit{StackedAttentionNet} \textbf{[SAN]} An adaptation of
a model devised by \citet{Yang2016} which was originally proposed for
VQA.
We replace our interpreter with their stacked attention model
to compute the attention map~$x_{\text{loc}}$.
Instead of employing a pretrained CNN as they did, we train a CNN
from scratch to accommodate to \textsc{xworld}.
The CNN is configured as the one employed by our model.
The rest of our model is unchanged.

\noindent\textit{VIS-LSTM} \textbf{[VL]} An adaptation of a model
devised by \citet{Ren2015} which was originally proposed for VQA.
We flatten~$h$ and project it to the word embedding
space~$\mathbb{R}^D$.
Then it is appended to the input sentence~$s$ as the first word.
The augmented sentence goes through an LSTM whose last state is used
for both NAV and QA (Figure~\ref{fig:VL}, Appendix~\ref{app:baselines}).

\noindent\textit{ConcatEmbed} \textbf{[CE]} An adaptation of a model
proposed by \citet{Mao2015} which was originally proposed for image
captioning.
It instantiates~$\mathbf{L}$ as a vanilla LSTM which outputs a
sentence embedding.
Then~$h$ is projected and concatenated with the embedding.
The concatenated vector is used for both NAV and QA
(Figure~\ref{fig:CE} Appendix~\ref{app:baselines}).
This concatenation mechanism is also employed
by~\citet{Hermann17,Misra17}.

\noindent \textit{NavAlone} \textbf{[NAVA]} An ablation of our model
that does not have the QA pipeline ($\mathbf{M}_P$ and $\mathbf{P}$)
and is trained only on the NAV tasks.
The rest of the model is the same.

In the following experiments, we train all six approaches (four
comparison methods, one ablation, and our model) with a small learning
rate of $1\times 10^{-5}$ and a batch size of 16, for a maximum of
200k minibatches.
Additional training details are described in
Appendix~\ref{app:details}.
After training, we test each approach on 50k sessions.
For NAV, we compute the average success rate of navigation where a
success is defined as reaching the correct location before the time
out of a session.
For QA, we compute the average accuracy in answering the questions.

\begin{figure}[t]
  \begin{center}
    \begin{tabular}{@{}c@{}}
      \begin{tabular}{@{}c@{}}
        \includegraphics[width=0.70\textwidth]{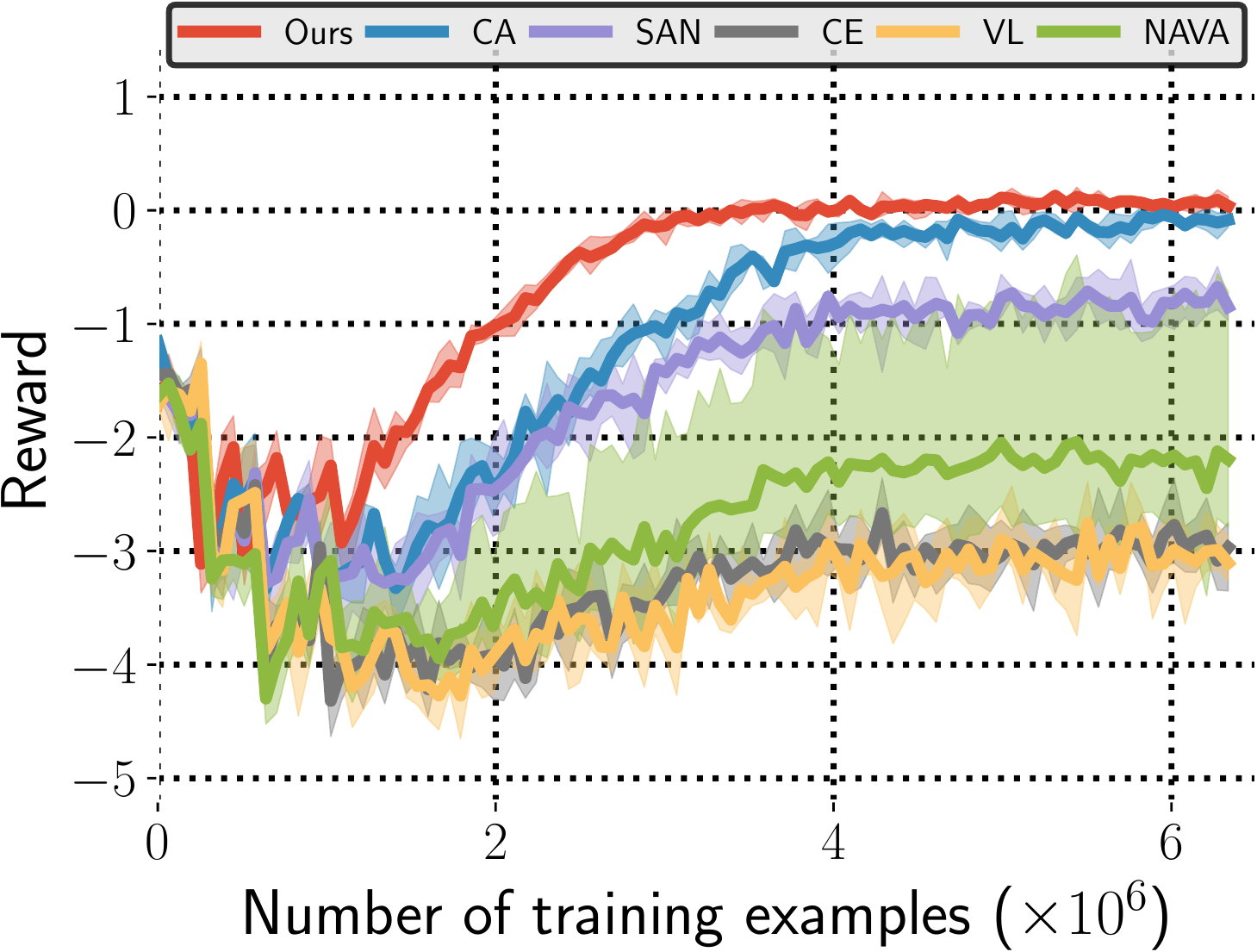}\\
        (a)\\
      \end{tabular}\\
      \begin{tabular}{@{}cc@{}}
        \includegraphics[width=0.66\textwidth]{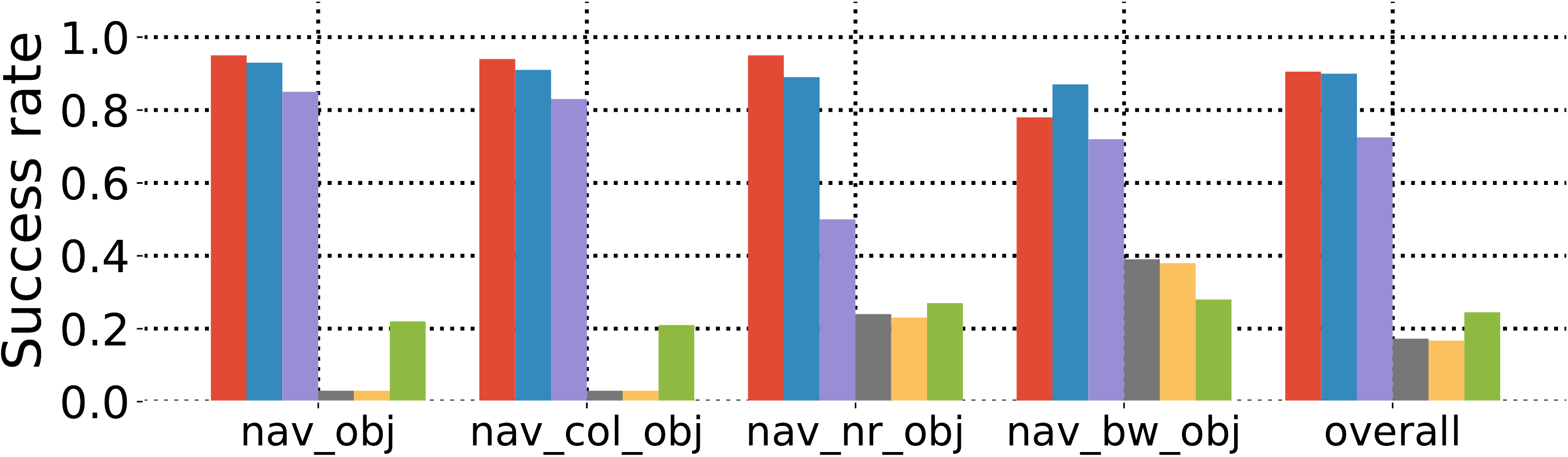}
        &
        \raisebox{6pt}{\includegraphics[width=0.29\textwidth]{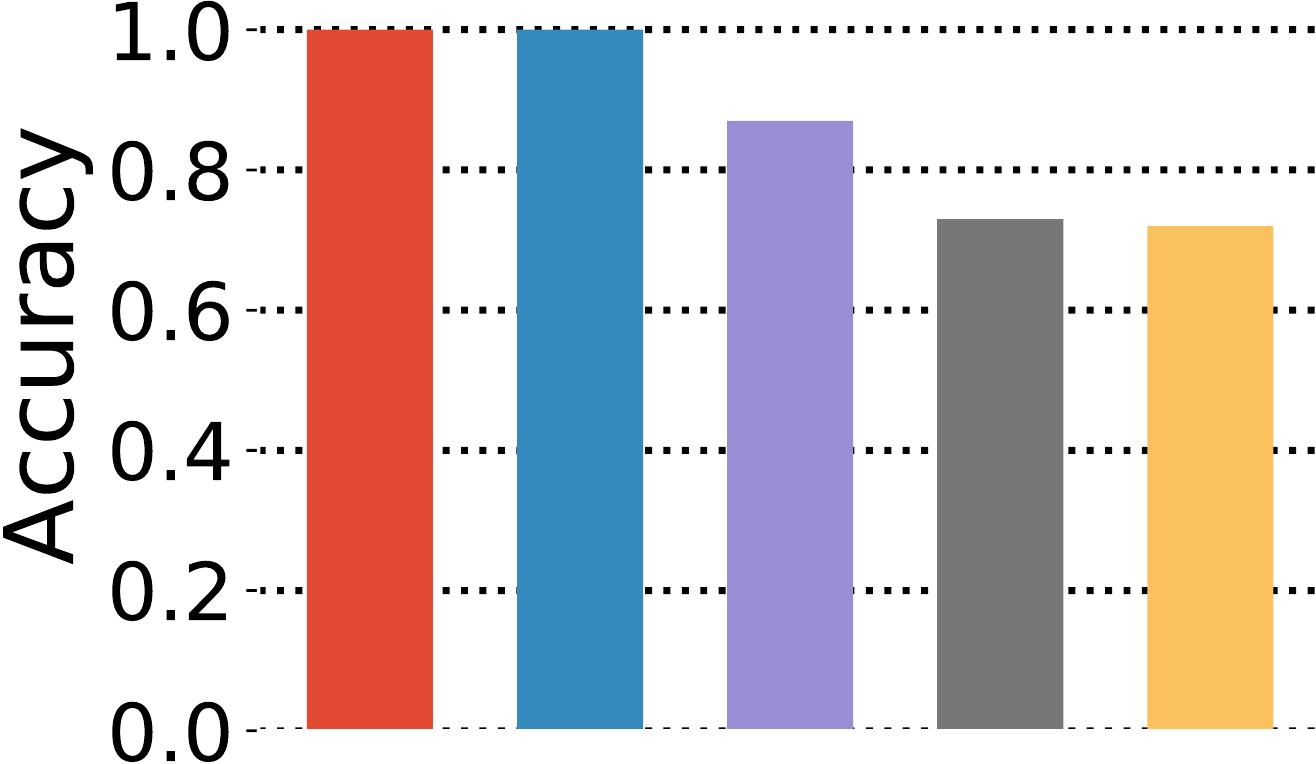}}\\
        (b) & (c)\\
      \end{tabular}\\\\
    \end{tabular}
    \caption{The basic evaluation.
      (a) Training reward curves.
      The shown reward is the accumulated discounted reward per
      session, averaged over every 8k training time steps.
      The shaded area of each curve denotes the variance among 4
      random initializations of the model parameters.
      The reason why the curves tend to drop in the beginning is
      that the map difficulty increases according to our curriculum
      (Appendix~\ref{app:curriculum}).
      (b) Navigation success rates in the test.
      (c) The accuracies of the answers in the test (\textbf{NAVA} is
      excluded because it does not train QA).
    }
    \label{fig:basic-eval}
  \end{center}
\end{figure}

\subsection{Basic evaluation}
\label{sec:baselines}
In this experiment, the training and testing sentences (including NAV
commands and QA questions) are sampled from the same distribution over
the entire sentence space.
We call it the normal language setting.\footnote{Although some test
  sentences might not be seen in training (\ie\ zero-shot) due to
  sampling, all the content words and their combinations (totaling a
  dozen thousands) are highly likely to be exhausted by training.
  Thus we consider this experiment as a normal setting, compared to
  the zero-shot setting in Section~\ref{sec:zero-shot}.
}

The training reward curves and testing results are shown in
Figure~\ref{fig:basic-eval}.
\textbf{VL} and \textbf{CE} have quite low rewards without
convergences.
These two approaches do not use the spatial attention~$x_{\text{loc}}$,
and thus always attend to whole images with no focus.
The region of a target location is tiny compared to the whole
egocentric image (a ratio of $1:(7\times2-1)^2=1:169$).
It is possible that this difficulty drives the models towards
overfitting QA without learning useful representations for NAV.
Both~\textbf{CA} and~\textbf{SAN} obtain rewards and success rates
slightly worse than ours.
This suggests that in a normal language setting, existing attention
models are able to handle previously seen sentences.
However, their language generalization abilities, especially on the
ZS2 sentences, are usually very weak, as we will demonstrate in the
next section.
The ablation~\textbf{NAVA} has a very large variance in its
performance.
Depending on the random seed, its reward can reach that of
\textbf{SAN} ($-0.8$), or it can be as low as that of \textbf{CE}
($-3.0$).
Comparing it to our full method, we conclude that even though the QA
pipeline operates on a completely different set of sentences, it
learns useful local sentential knowledge that results in an effective
training of the NAV pipeline.

Note that all the four comparison methods obtained high QA accuracies
($>$70\%, see Figure~\ref{fig:basic-eval}c), despite their distinct NAV
results.
This suggests that QA, as a supervised learning task, is easier than
NAV as an RL task in our scenario.
One reason is that the groundtruth label in QA is a much stronger
training signal than the reward in NAV.
Another reason might be that NAV additionally requires learning the
control module, which is absent in QA.

\begin{figure}[t]
  \begin{center}
    \resizebox{0.98\textwidth}{!}{
      \begin{tabular}{@{}c@{}}
        \includegraphics[width=\textwidth]{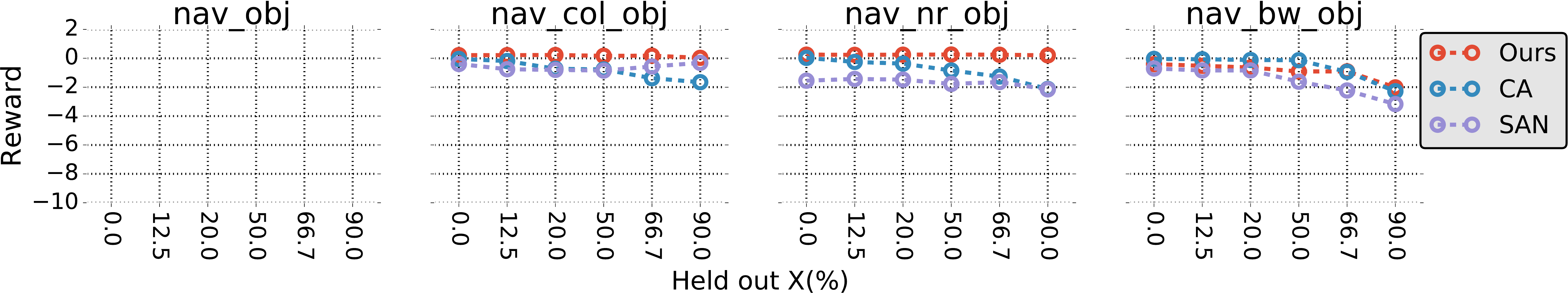}\\
        (a)\\
        \includegraphics[width=\textwidth]{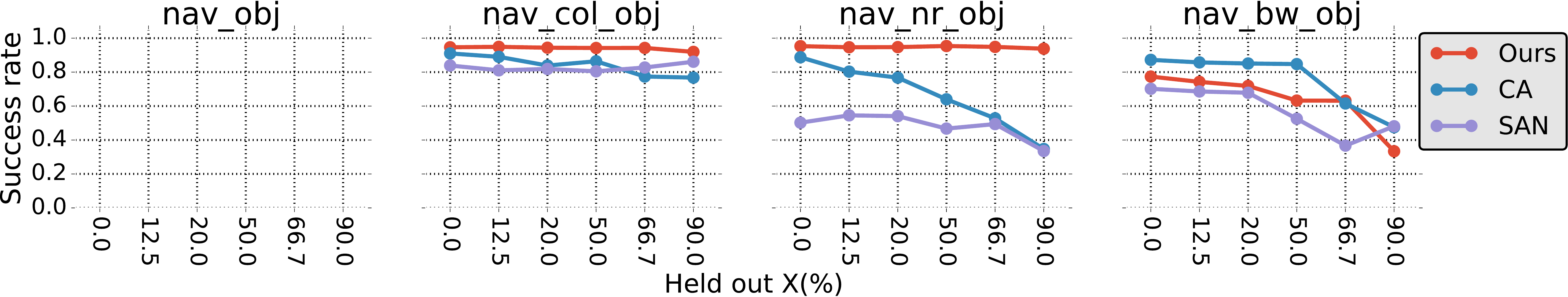}\\
        (b)\\
        \includegraphics[width=0.34\textwidth]{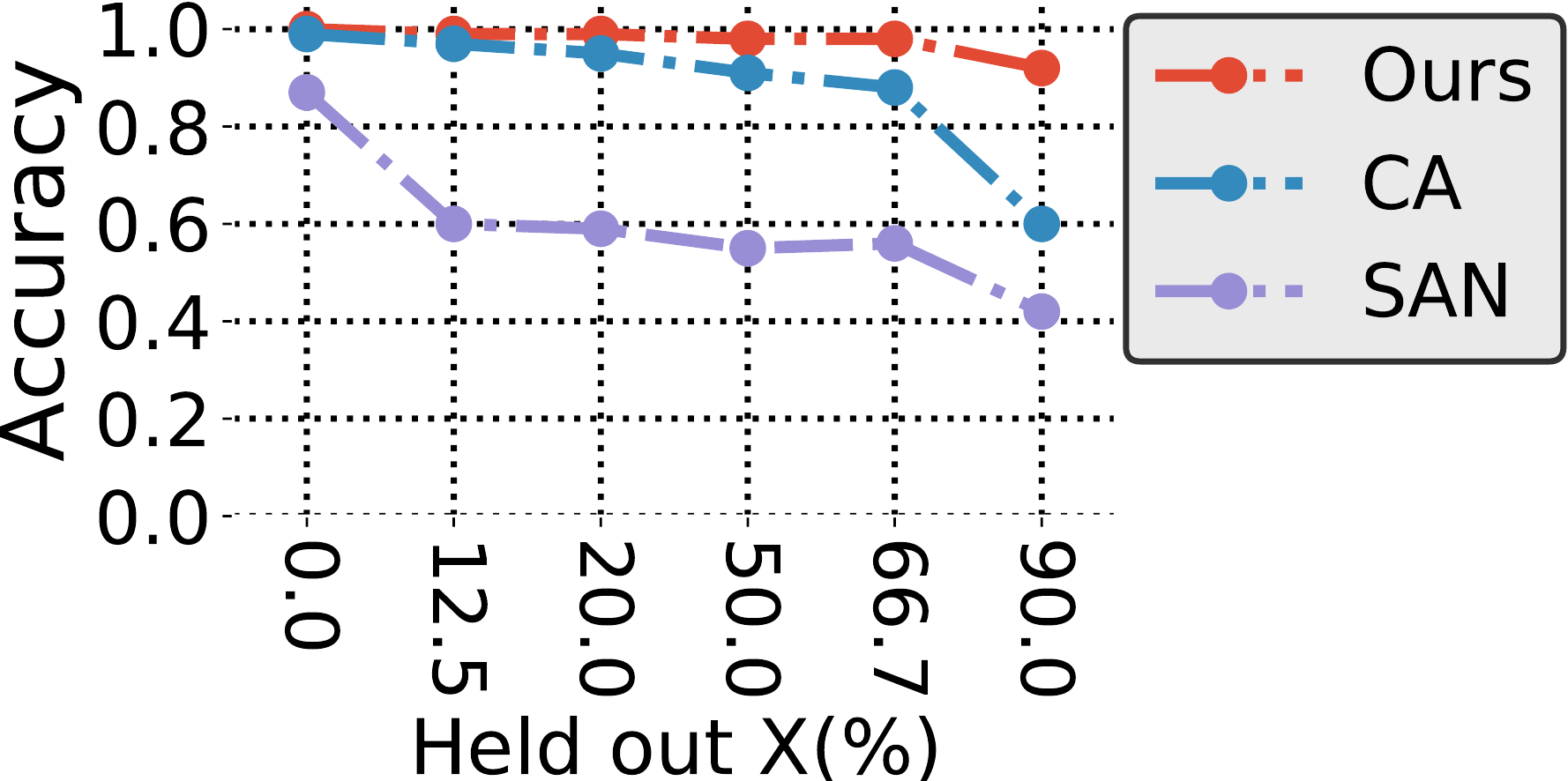}\\
        (c)\\
        \includegraphics[width=\textwidth]{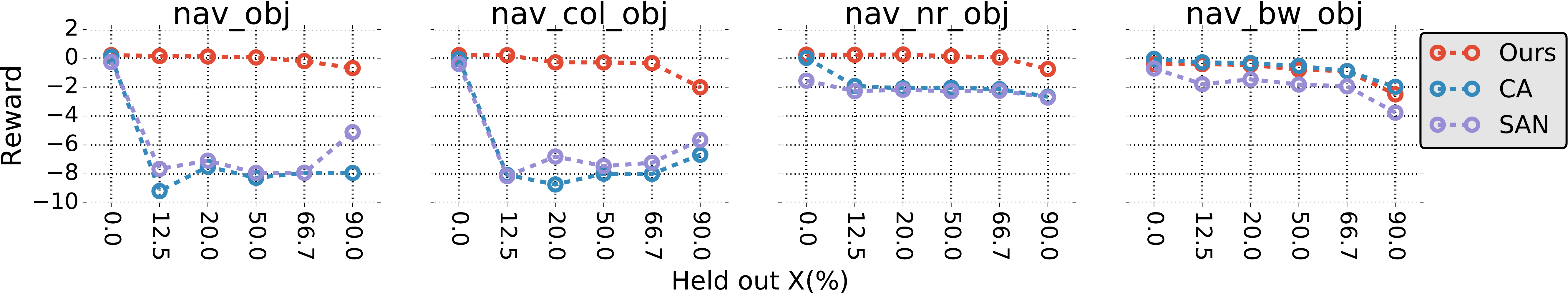}\\
        (d)\\
        \includegraphics[width=\textwidth]{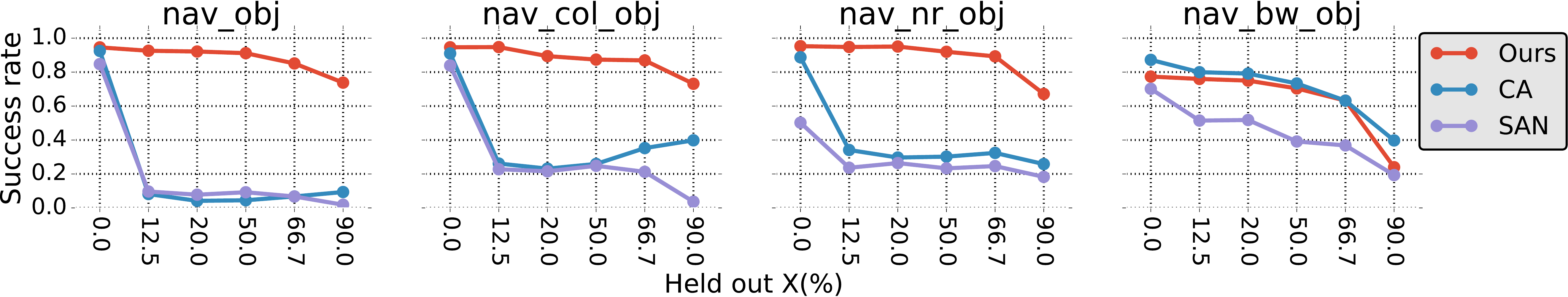}\\
        (e)\\
        \includegraphics[width=0.34\textwidth]{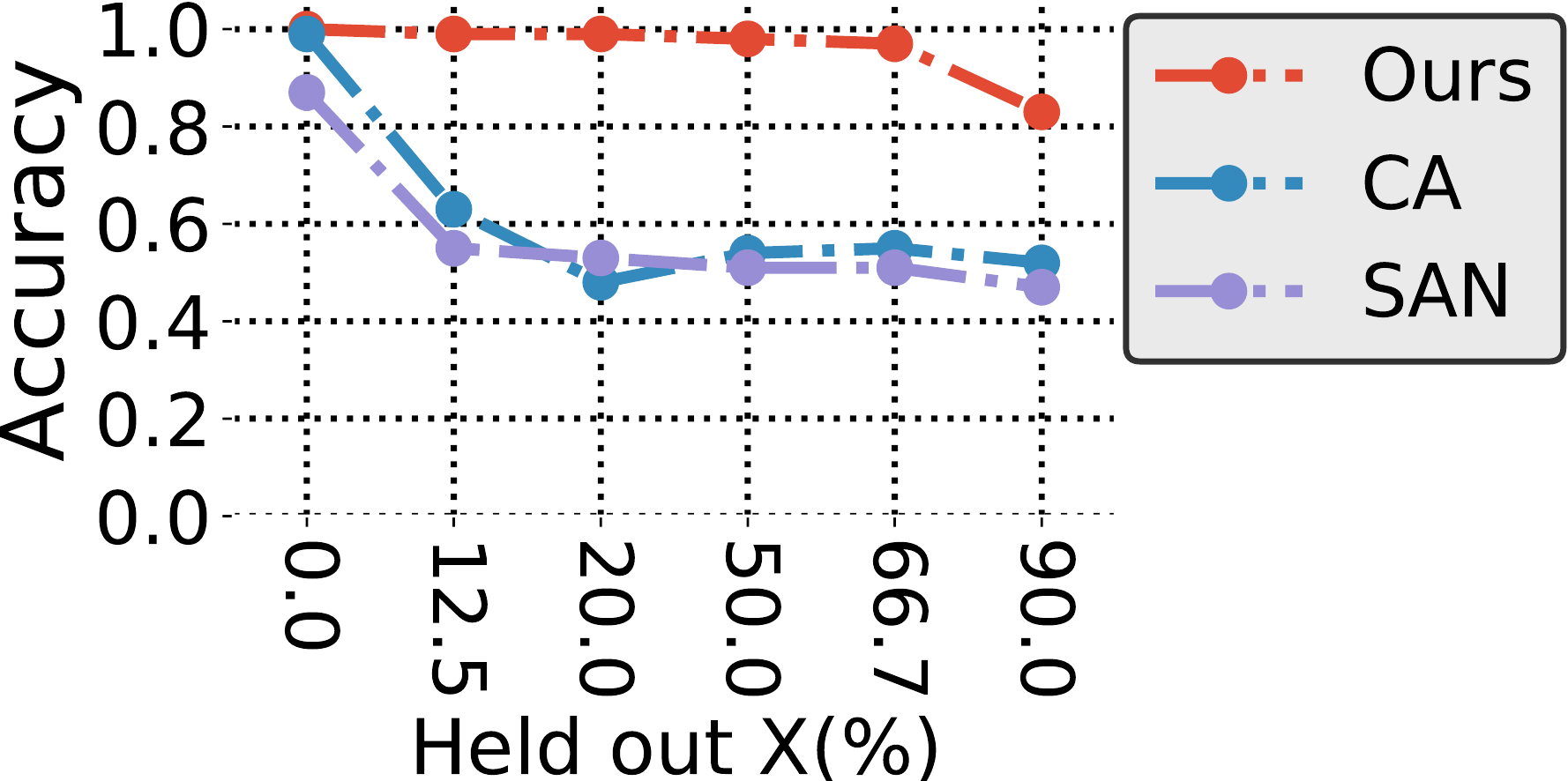}\\
        (f)\\
      \end{tabular}
    }
      \caption{The test results of language generalization with a varying
        held-out portion of $X$\%, where $X=0$ represents the basic
        evaluation in Section~\ref{sec:baselines}.
        (a--c) \textbf{ZS1}.
        (d--f) \textbf{ZS2}.
        For either \textbf{ZS1} or \textbf{ZS2}, from top to bottom,
        the three rows represent the average navigation reward per session,
        the average navigation success rate per session, and the
        average QA accuracy, respectively.
        (The plots of \texttt{nav\_obj} in (a) and (b) are empty
        because there is no ZS1 sentence of this type by definition.)
      }
      \label{fig:zero-shot-res}
    \end{center}
\end{figure}

\subsection{Language generalization}
\label{sec:zero-shot}
Our more important question is whether the agent has the ability of
interpreting zero-shot sentences.
For comparison, we use \textbf{CA} and~\textbf{SAN} from the previous
section, as they achieved good performance in the normal language
setting.
For a zero-shot setting, we set up two language conditions:

\noindent\textit{NewWordCombination} \textbf{[ZS1]} Some word pairs
are excluded from both the NAV commands and the QA questions.
We consider three types of unordered combinations of the content
words: (\textit{object}, \textit{spatial relation}), (\textit{object},
\textit{color}), and (\textit{object}, \textit{object}).
We randomly hold out $X$\% of the word pairs during the training.

\noindent\textit{NewWord} \textbf{[ZS2]} Some single words are
excluded from both the NAV commands and the QA questions, but can
appear in the QA answers.
We randomly hold out $X$\% of the object words during the training.

We vary the value of~$X$ (12.5, 20.0, 50.0, 66.7, or 90.0) in both
conditions to test how sensitive the generalization is to the amount
of the held-out data.
For evaluation, we report the test results only for the zero-shot
sentences that contain the held-out word pairs or words.
The results are shown in Figure~\ref{fig:zero-shot-res}.

We draw three conclusions from the results.
First, the ZS1 sentences are much easier to interpret than the ZS2
sentences.
Neural networks seem to inherently have this capability to some
extent.
This is consistent with what has been observed in some previous
work~\citep{Hermann17,Chaplot18} that addresses the generalization
on new word combinations.
Second, the ZS2 sentences are difficult for \textbf{CA} and \textbf{SAN}.
Even with a held-out portion as small as $X\%=12.5\%$, their navigation
success rates and QA accuracies drop up to 80\% and 35\%, respectively,
compared to those in the normal language setting.
In contrast, our model tends to maintain the same results until
$X=90.0$.
Impressively, it achieves an average success rate of 60\% and an
average accuracy of 83\% even when the number of new object words is 9
times that of seen object words in the NAV commands and QA questions,
respectively!
Third, in \textbf{ZS2}, if we compare the slopes of the success-rate
curves with those of the accuracy curves (as shown in
Figures~\ref{fig:zero-shot-res}e and~\ref{fig:zero-shot-res}f), we
notice that the agent generalizes better on QA than on NAV.
This further verifies our finding in the previous section that QA is
in general an easier task than NAV in \textsc{xworld}.
This demonstrates the necessity of evaluating NAV in addition to QA, as
NAV requires additional language grounding to control.

Interestingly, we notice that \texttt{nav\_bw\_obj} is an outlier
command type for which \textbf{CA} is much less sensitive to the
increase of $X$ in \textbf{ZS2}.
A deep analysis reveals that for \texttt{nav\_bw\_obj}, \textbf{CA}
learns to cheat by looking for the image region that contains the
special pattern of object pairs in the image without having to
recognize the objects.
This suggests that neural networks tend to exploit data in an
unexpected way to achieve tasks if no constraints are imposed
\citep{Kottur2017}.

To sum up, our model exhibits a strong generalization ability on both
ZS1 and ZS2 sentences, the latter of which pose a great challenge for
traditional language grounding models.
Although we use a particular 2D world for evaluation in this work, the
promising results imply the potential for scaling to an even larger
vocabulary and grammar with a much larger language space.

\subsection{How does it adapt to 3D?}
We discuss the possibility of adapting our model to an agent with
similar language abilities in a 3D world
(\eg\ \citet{Beattie2016,Johnson2016}).
This is our goal for the future, but here we would like to share some
preliminary thoughts.
Generally speaking, a 3D world will pose a greater challenge for
vision-related computations.
The key element of our model is the attention cube~$x_{\text{cube}}$
that is intended for an explicit language grounding, including the
channel mask~$x_{\text{feat}}$ and the attention map~$x_{\text{loc}}$.
The channel mask only depends on the sentence, and thus is expected to
work regardless of the world's dimensionality.
The interpreter depends on a sequence of score maps~$\chi$ which for
now are computed as multiplying a word embedding with the feature cube
(Eq.~\ref{eq:phi}).
A more sophisticated definition of~$\phi$ will be needed to detect
objects in a 3D environment.
Additionally, the interpreter models the spatial transform of
attention as a 2D convolution (Eq.~\ref{eq:aggregation}).
This assumption will be not valid for reasoning 3D spatial relations
on 2D images, and a better transform method that accounts for
perspective distortion is required.
Lastly, the surrounding environment is only partially observable to
a 3D agent.
A working memory, such as an LSTM added to the action
module~$\mathbf{A}$, will be important for navigation in this case.
Despite these changes to be made, we believe that our general explicit
grounding strategy and the common detection function shared by
language grounding and prediction have shed some light on the
adaptation.

\section{Conclusion}
We have presented an end-to-end model of a virtual agent for acquiring
language from a 2D world in an interactive manner, through the visual
and linguistic perception channels.
After learning, the agent is able to both interpolate and extrapolate
to interpret zero-shot sentences that contain new word combinations or
even new words.
This generalization ability is supported by an explicit grounding
strategy that disentangles the language grounding from the subsequent
language-independent computations.
It also depends on sharing a detection function between the language
grounding and prediction as the core computation.
This function enables the word meanings to transfer from the
prediction to the grounding during the test time.
Promising language acquisition and generalization results have been
obtained in the 2D \textsc{xworld}.
We hope that this work can shed some light on acquiring and
generalizing language in a similar way in a 3D world.

\subsubsection*{Acknowledgments}
We thank the anonymous reviewers for providing valuable comments and
suggestions.
We thank the other team members, Yuanpeng Li, Liang Zhao, Yi Yang, Zihang Dai, Qing Sun,
Jianyu Wang, and Xiaochen Lian, for helpful discussions.
We thank Jack Feerick and Richard Mark for proofreading.
Finally, we specially thank Dhruv Batra for his feedback on an early
version of this paper.

\bibliography{iclr2018}
\bibliographystyle{iclr2018}

\include{iclr2018_appendix}

\end{document}

%% file: iclr2018_appendix.tex
\appendix

\section*{\textbf{Appendices}}

\begin{figure}[t]
  \begin{center}
    \includegraphics[width=\textwidth]{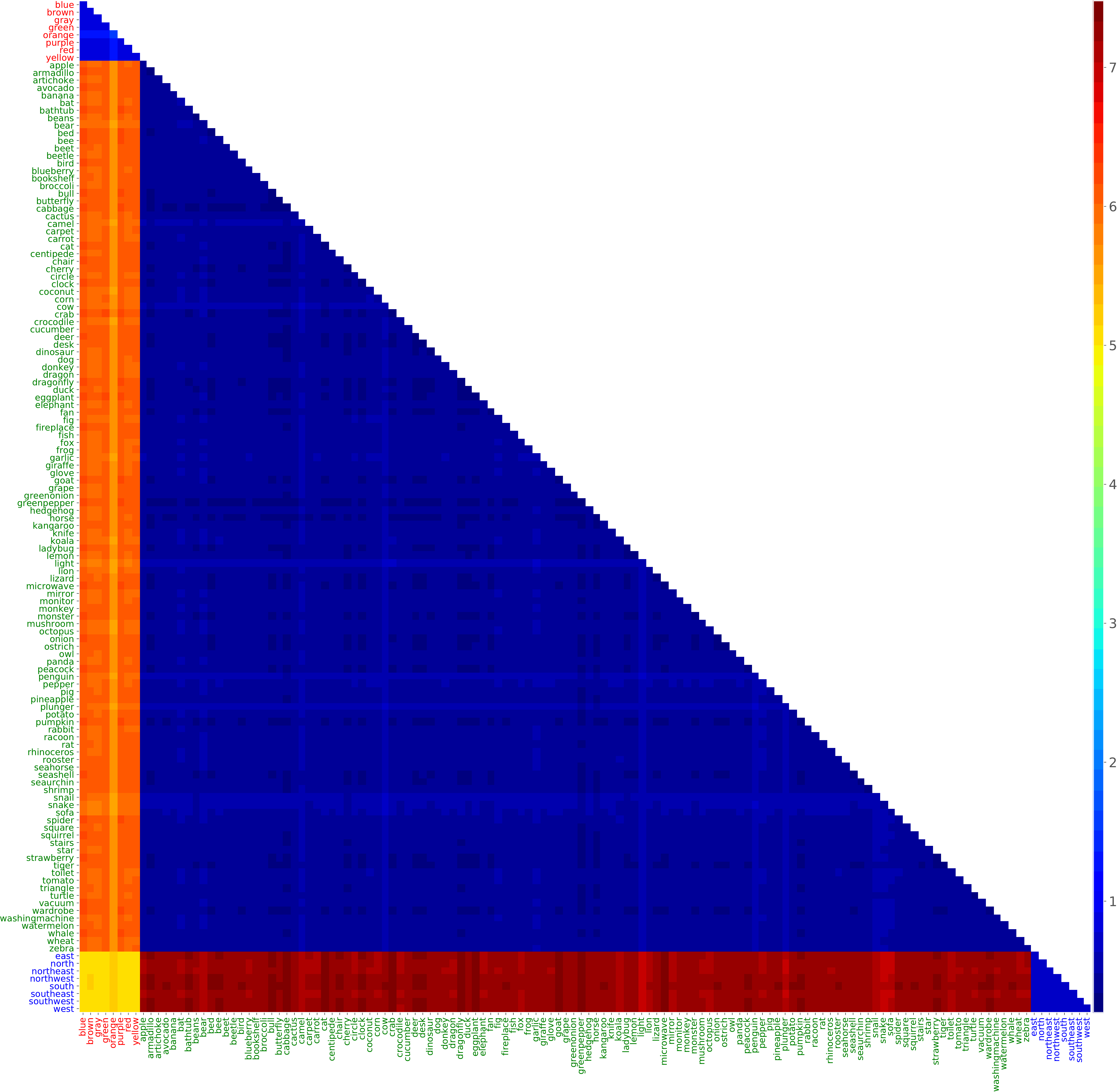}
    \caption{The Euclidean distance matrix of the 134 question groups
      where each group is represented by a word label.
      Each row (column) represents the sampled questions that
      have the word label as the answer.
      A matrix entry indicates the empirical expectation of the
      distance between the channel masks of the sentences from two
      question groups.
      The labels are arranged into three topics: {\color{red}
      color}, {\color{myGreen} object}, and {\color{blue} spatial
      relation}.
      A small distance indicates that the two channel masks are
      similar.
      (Zoom in on the screen for a better view.)
    }
    \label{fig:vis-x-feat}
  \end{center}
\end{figure}

\section{Visualization and analysis}
\label{sec:visualize}
In this section, we visualize and analyze some intermediate results of
our trained model.

\noindent\textbf{Channel mask}~$x_{\text{feat}}$\textbf{.} We inspect the channel
mask~$x_{\text{feat}}$ which allows the model to select certain
feature maps from a feature cube~$h$ and predict an answer to the
question~$s$.
We randomly sample 10k QA questions and compute~$x_{\text{feat}}$ for each of
them using the grounding module~$\mathbf{L}$.
We divide the 10k questions into 134 groups, where each group
corresponds to a different answer.\footnote{The word ``orange'' is both
a color word and an object word, which is why the number of groups is
one less than 119 (objects) $+$ 8 (spatial relations without
``between'') $+$ 8 (colors) = 135.}
Then we compute an Euclidean distance matrix~$D$ where entry $D[i,j]$
is the average distance between the~$x_{\text{feat}}$ of a question from the
$i$th group and that from the $j$th group
(Figure~\ref{fig:vis-x-feat}).
It is clear that there are three topics (object, color, and spatial
relation) in the matrix.
The distances computed within a topic are much smaller than
those computed across topics.
This demonstrates that with the channel mask, the model is able to
look at different subsets of features for questions of different
topics, while using the same subset of features for questions of the
same topic.
It also implies that the feature cube~$h$ is learned to organize
feature maps according to image attributes.

\begin{figure}[t]
  \begin{center}
    \begin{tabular}{@{}cc@{}}
      \includegraphics[scale=0.37, valign=t]{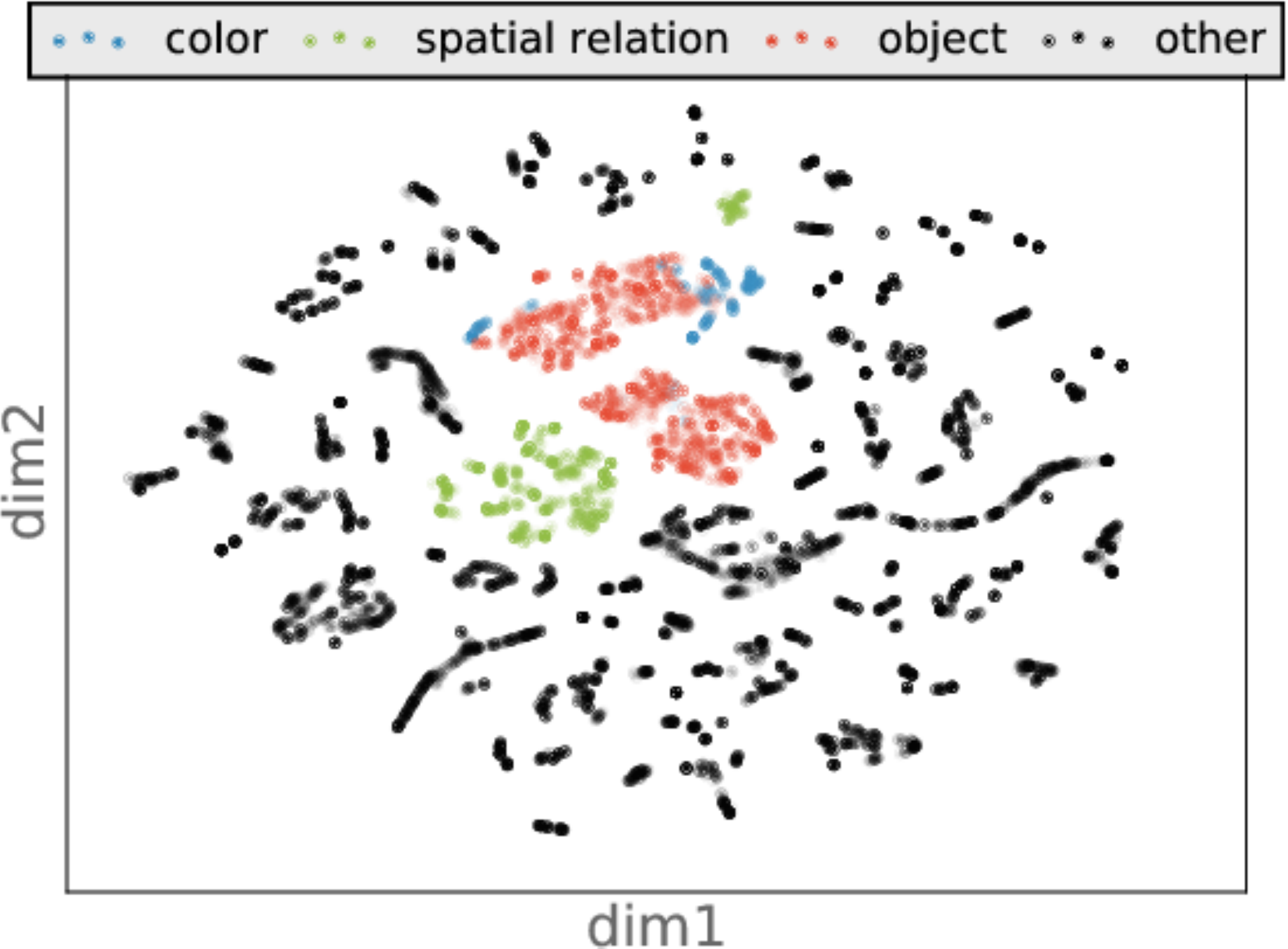}&
      \includegraphics[scale=0.4, valign=t]{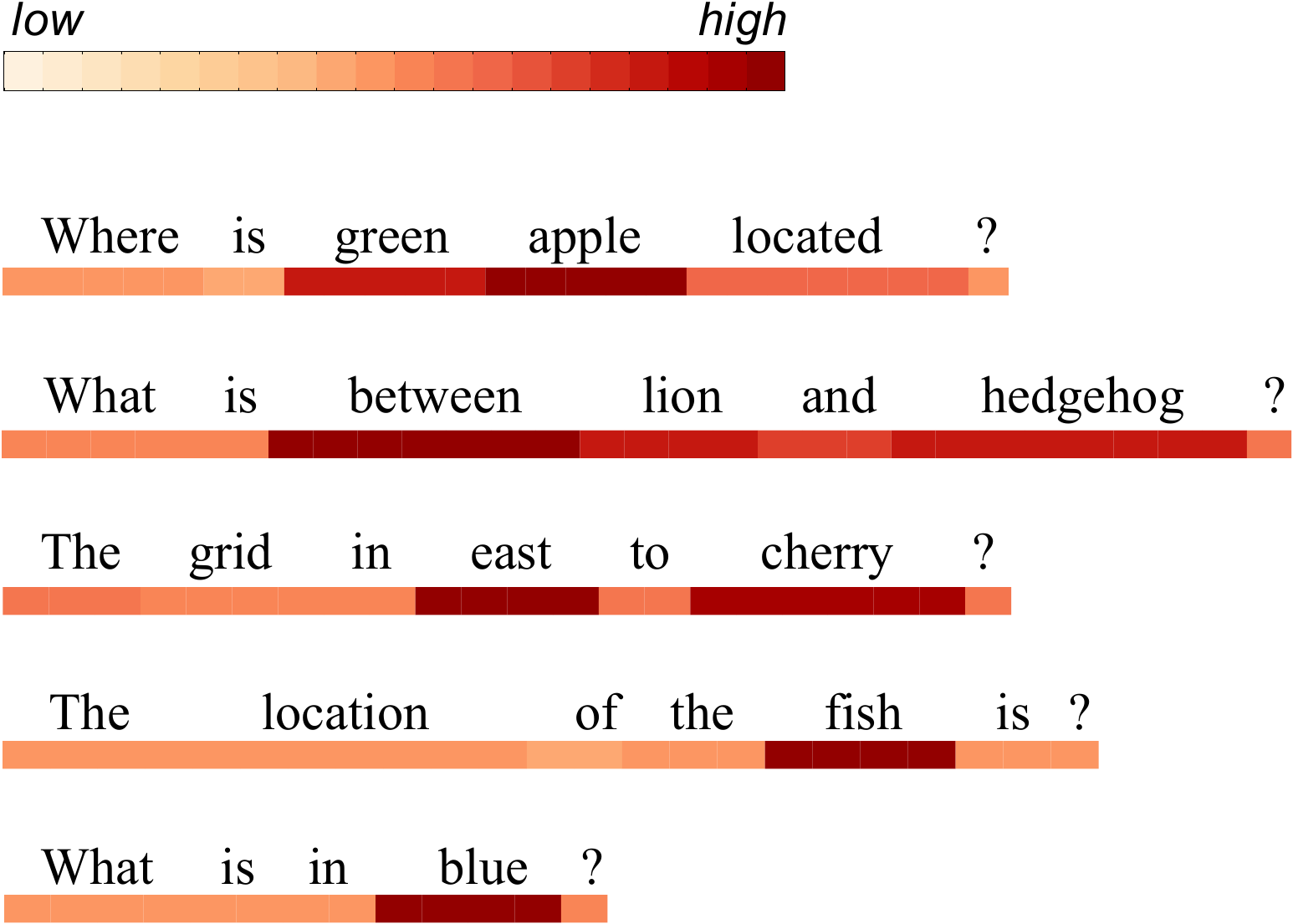}\\
      (a) & (b)\\
    \end{tabular}
    \caption{Visualizations of the computation of word attention.
      (a) Word context vectors~$\overline{w}_l$.
      (b) The word attentions~$o_l$ of several example questions.
      Each attention vector, represented by a color bar, shows the
      attention accumulated over~$I$ interpretation steps.
    }
    \label{fig:vis-interpreter}
  \end{center}
\end{figure}

\noindent\textbf{Word context \& attention.} To intuitively
demonstrate how the interpreter works, we visualize the word context
vectors~$\overline{w}_l$ in Eq.~\ref{eq:attention} for a total of 20k
word locations~$l$ (10k from QA questions and the other 10k from NAV
commands).
Each word context vector is projected down to a space of 50 dimensions
using PCA~\citep{Jolliffe86}, after which we use
t-SNE~\citep{Maaten08,Ulyanov2016} to further reduce the
dimensionality to 2.
The t-SNE method uses a perplexity of 100 and a learning rate of 200,
and runs for 1k iterations.
The visualization of the 20k points is shown in
Figure~\ref{fig:vis-interpreter}a.
Recall that in Eq.~\ref{eq:attention} the word attention is computed by
comparing the interpreter state~$p^{i-1}$ with the word context
vectors~$\overline{w}_l$.
In order to select the content words to generate meaningful score maps
via~$\phi$, the interpreter is expected to differentiate them from the
remaining grammatical words based on the contexts.
This expectation is verified by the above visualization in which the
context vectors of the content words (in blue, green, and red) and
those of the grammatical words (in black) are mostly separated.
Figure~\ref{fig:vis-interpreter}b shows some example questions
whose word attentions are computed from the word context vectors.
It can be seen that the content words are successfully selected by the
interpreter.

\noindent\textbf{Attention map}~$x_{\text{loc}}$\textbf{.} Finally, we visualize the
computation of the attention map~$x_{\text{loc}}$.
In each of the following six examples, the intermediate attention maps
$x_{\text{loc}}^i$ and word attentions $o^i_l$ (in Eq.~\ref{eq:aggregation})
of the three interpretation steps are shown from top to bottom.
Each step shows the current attention map $x_{\text{loc}}^i$ overlaid on the
environment image $e$.
The last attention map $x_{\text{loc}}^3$ is the final output of the
interpreter at the current time.
Note that not all the results of the three steps are needed to
generate the final output.
Some results might be discarded according to the value of the update
gate $\rho^i$.
As a result, sometimes the interpreter may produce ``bogus''
intermediate attention maps which do not contribute to
$x_{\text{loc}}$.
Each example also visualizes the environment terrain map
$x_{\text{terr}}$ (defined in Appendix~\ref{app:details}) that
perfectly detects all the objects and blocks, and thus provides a good
guide for the agent to navigate successfully without hitting walls or
confounding targets.
For a better visualization, the egocentric images are converted back
to the normal view.

\begin{figure}[t]
  \begin{center}
    \includegraphics[width=0.9\textwidth]{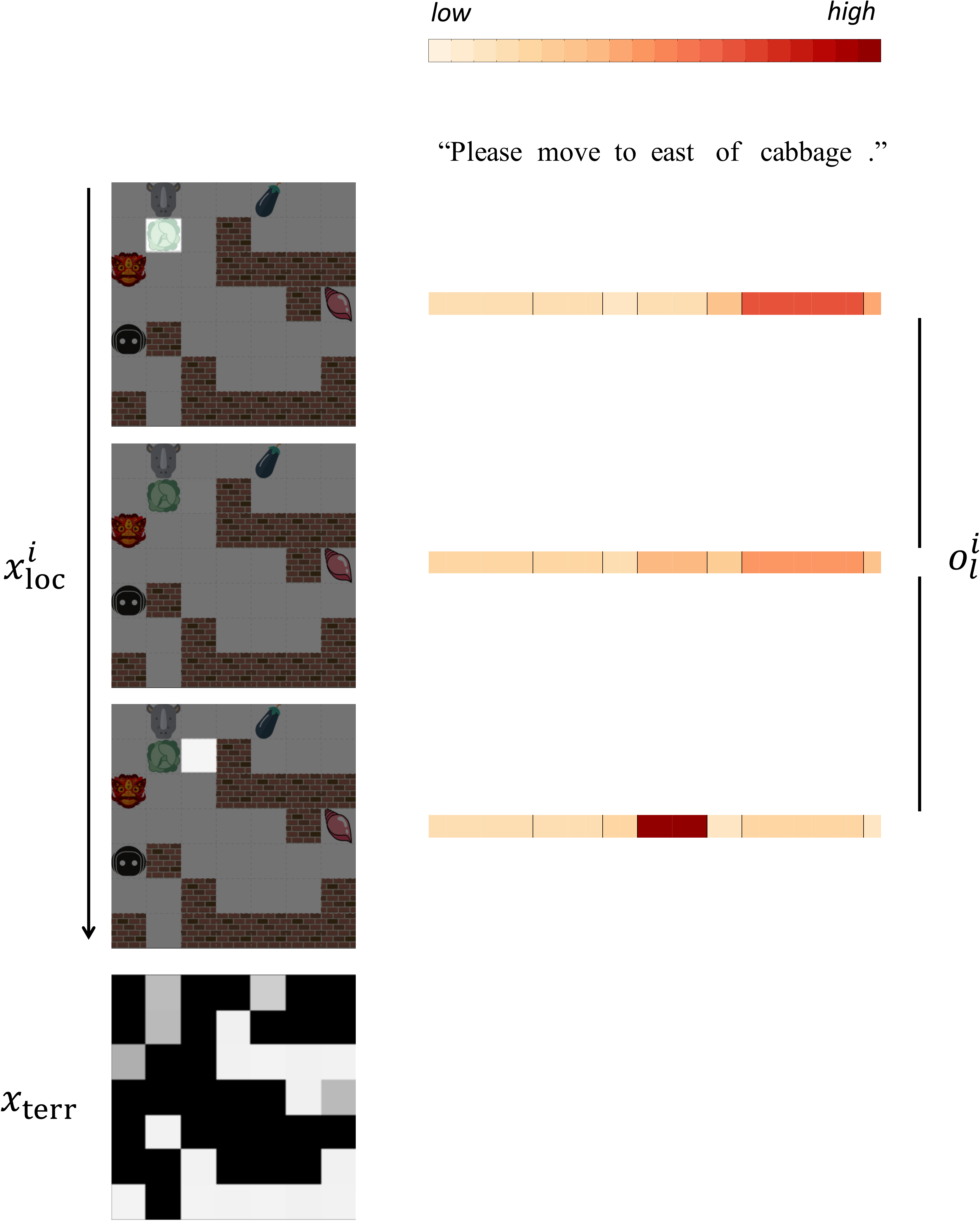}\\
  \end{center}
  \caption{The first example showing how $x_{\text{loc}}$ is computed.}
\end{figure}

\clearpage

\begin{figure}[t]
  \begin{center}
    \includegraphics[width=\textwidth]{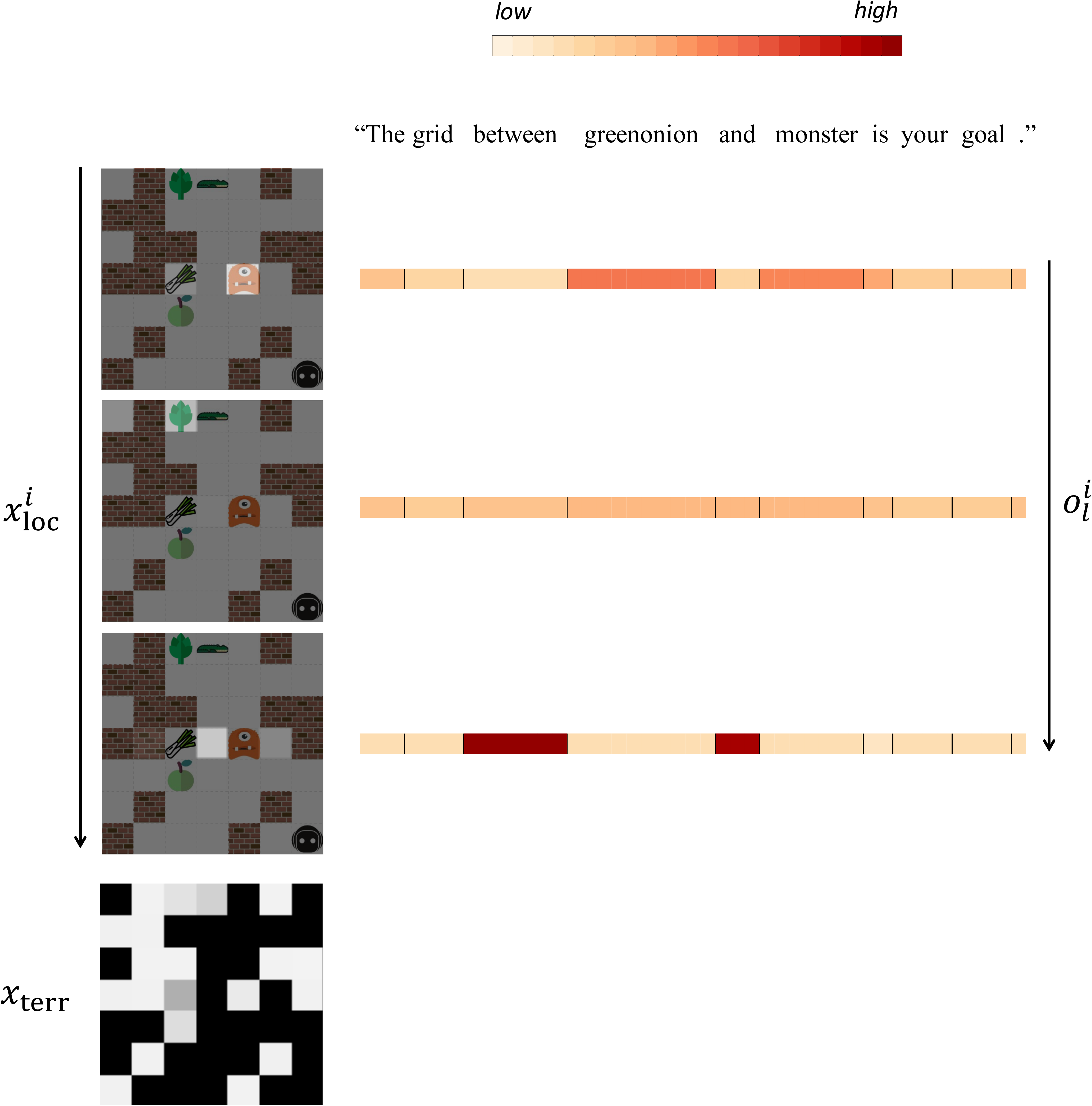}\\
  \end{center}
  \caption{The second example showing how $x_{\text{loc}}$ is computed.}
\end{figure}

\clearpage

\begin{figure}[t]
  \begin{center}
    \includegraphics[width=0.9\textwidth]{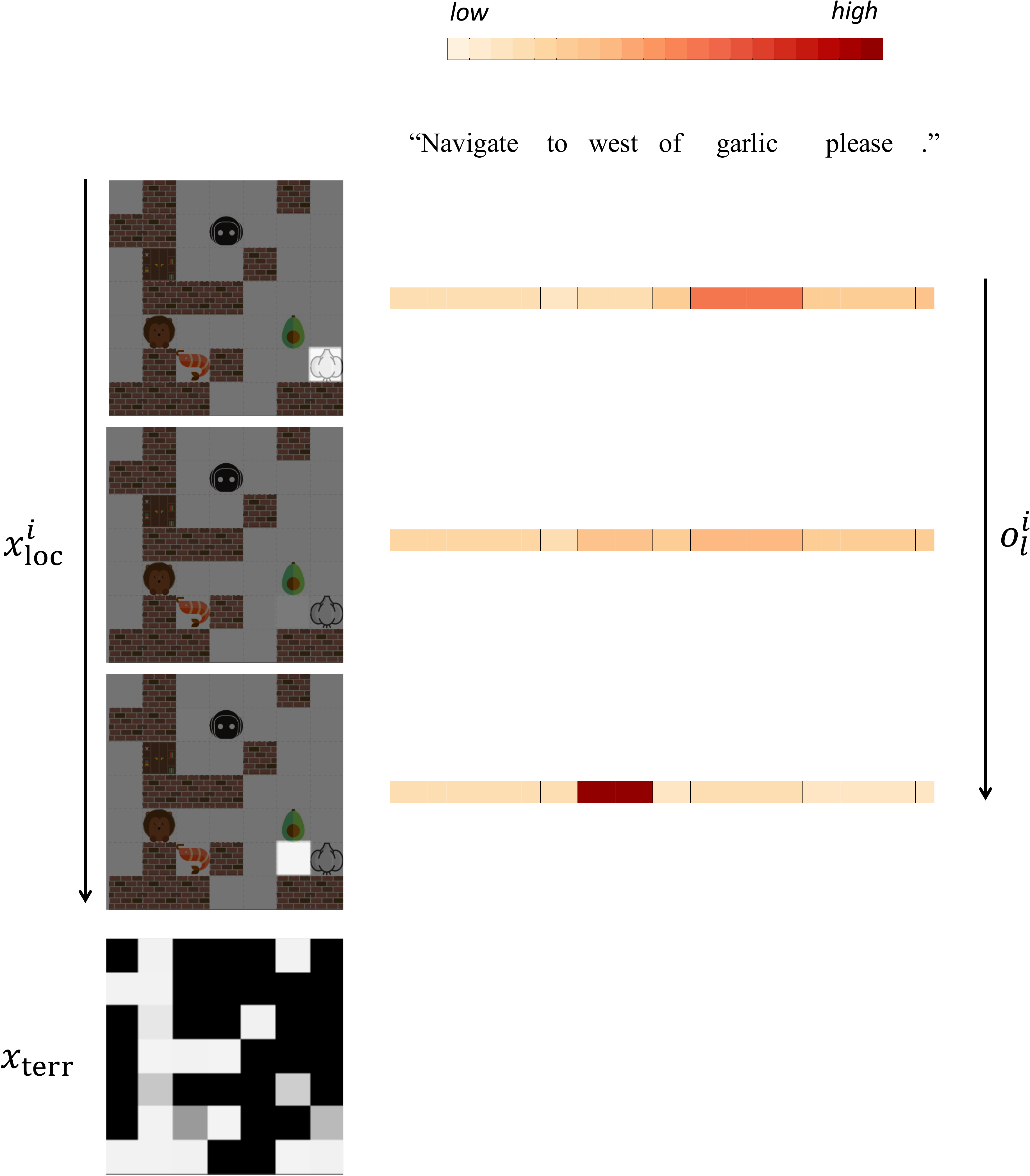}\\
  \end{center}
  \caption{The third example showing how $x_{\text{loc}}$ is computed.}
\end{figure}

\clearpage

\begin{figure}[t]
  \begin{center}
    \includegraphics[width=0.9\textwidth]{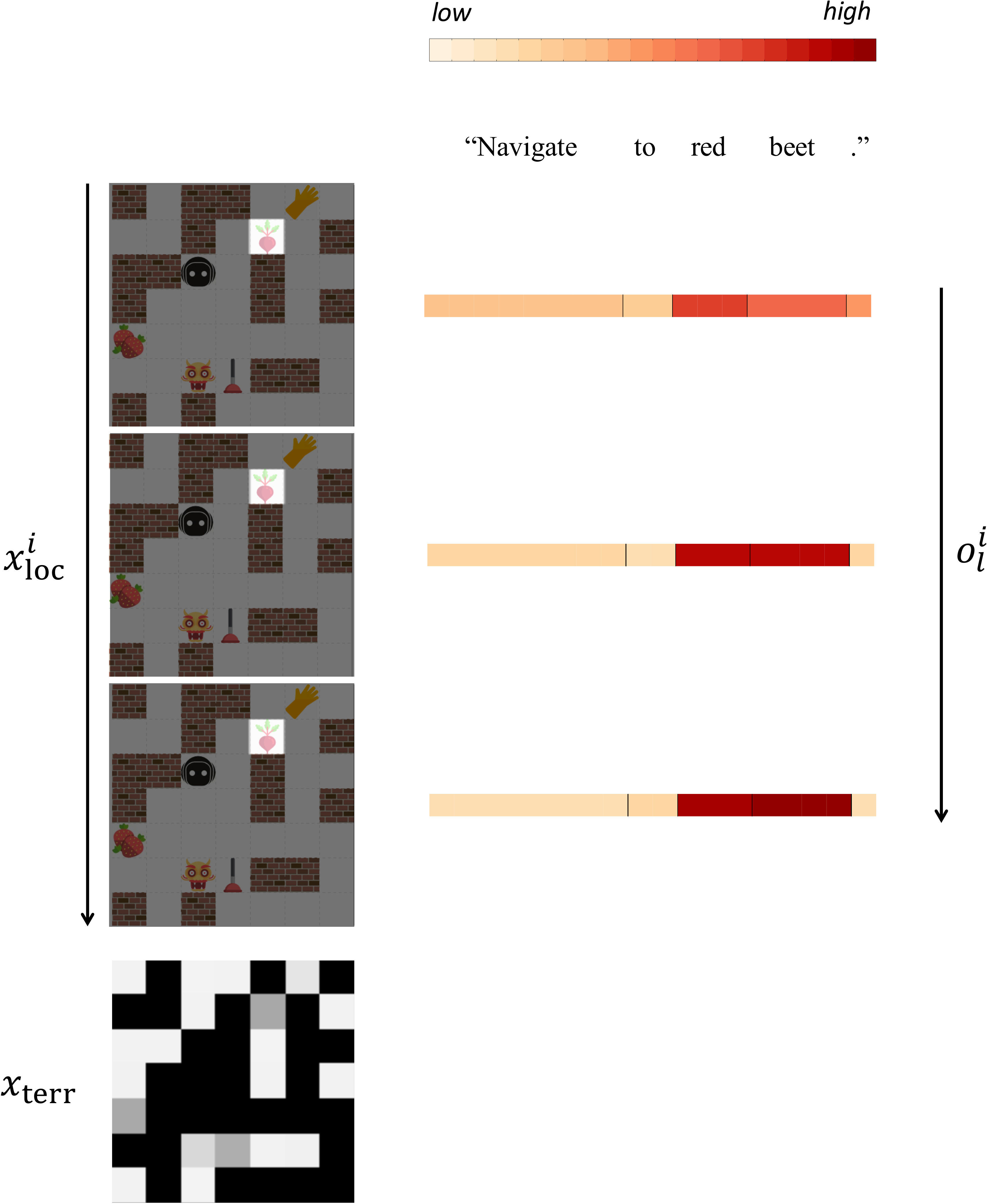}\\
  \end{center}
  \caption{The fourth example showing how $x_{\text{loc}}$ is computed.}
\end{figure}

\clearpage

\begin{figure}[t]
  \begin{center}
    \includegraphics[width=\textwidth]{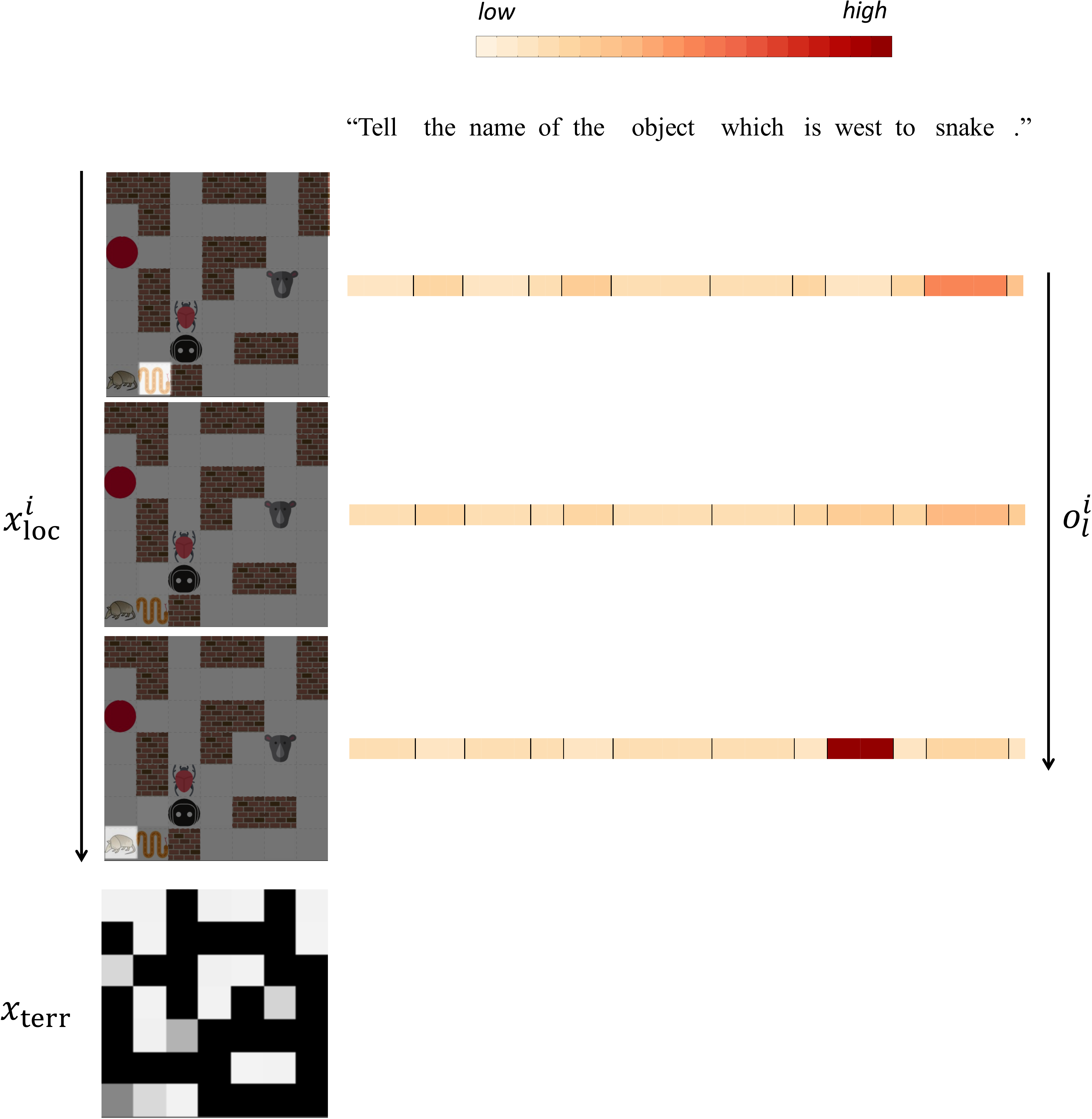}\\
  \end{center}
  \caption{The fifth example showing how $x_{\text{loc}}$ is computed.}
\end{figure}

\clearpage

\begin{figure}[t]
  \begin{center}
    \includegraphics[width=0.9\textwidth]{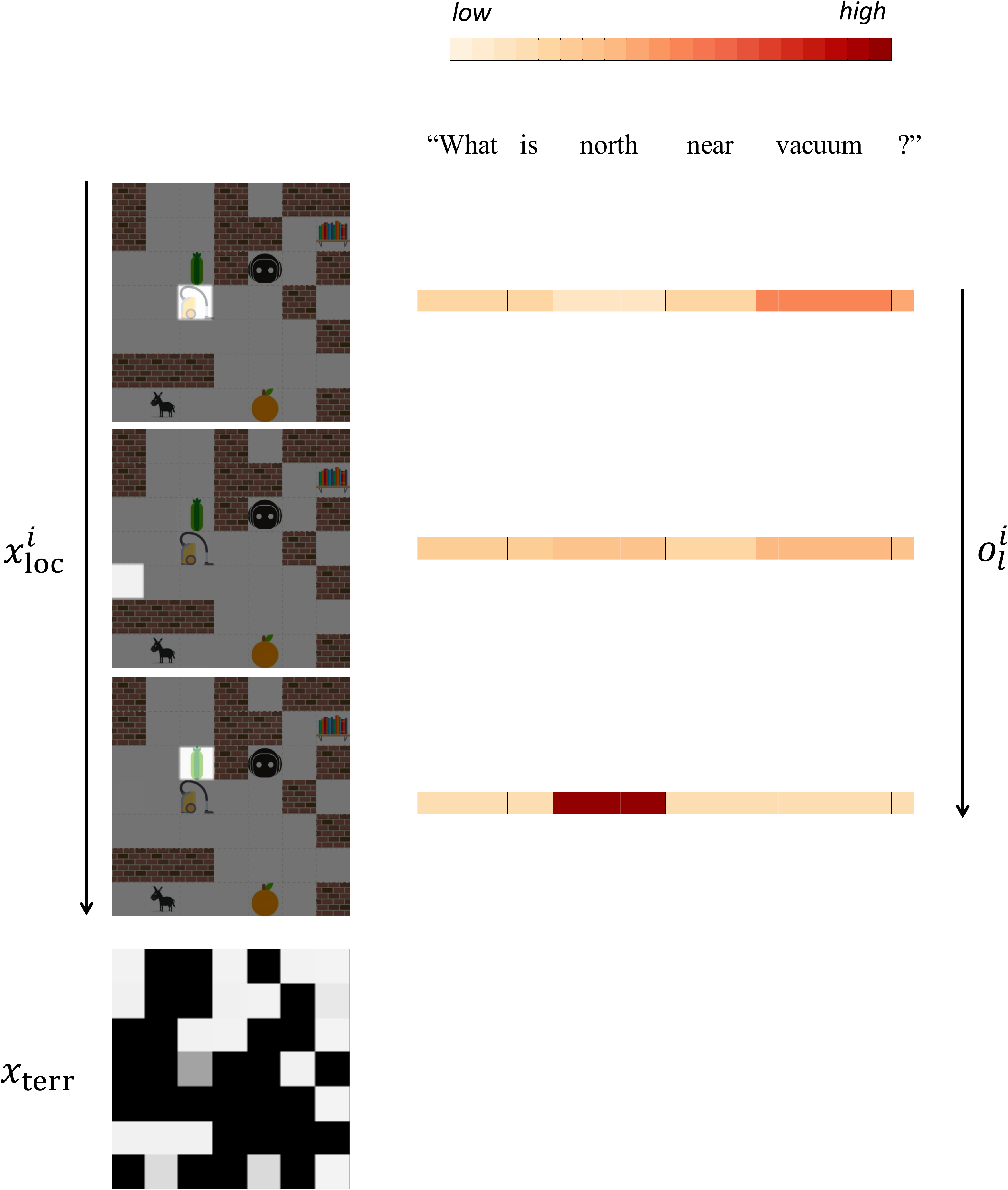}\\
  \end{center}
  \caption{The sixth example showing how $x_{\text{loc}}$ is computed.}
\end{figure}

\clearpage

\begin{figure}[t]
  \begin{center}
    \resizebox{\textwidth}{!}{
      \includegraphics[width=\textwidth]{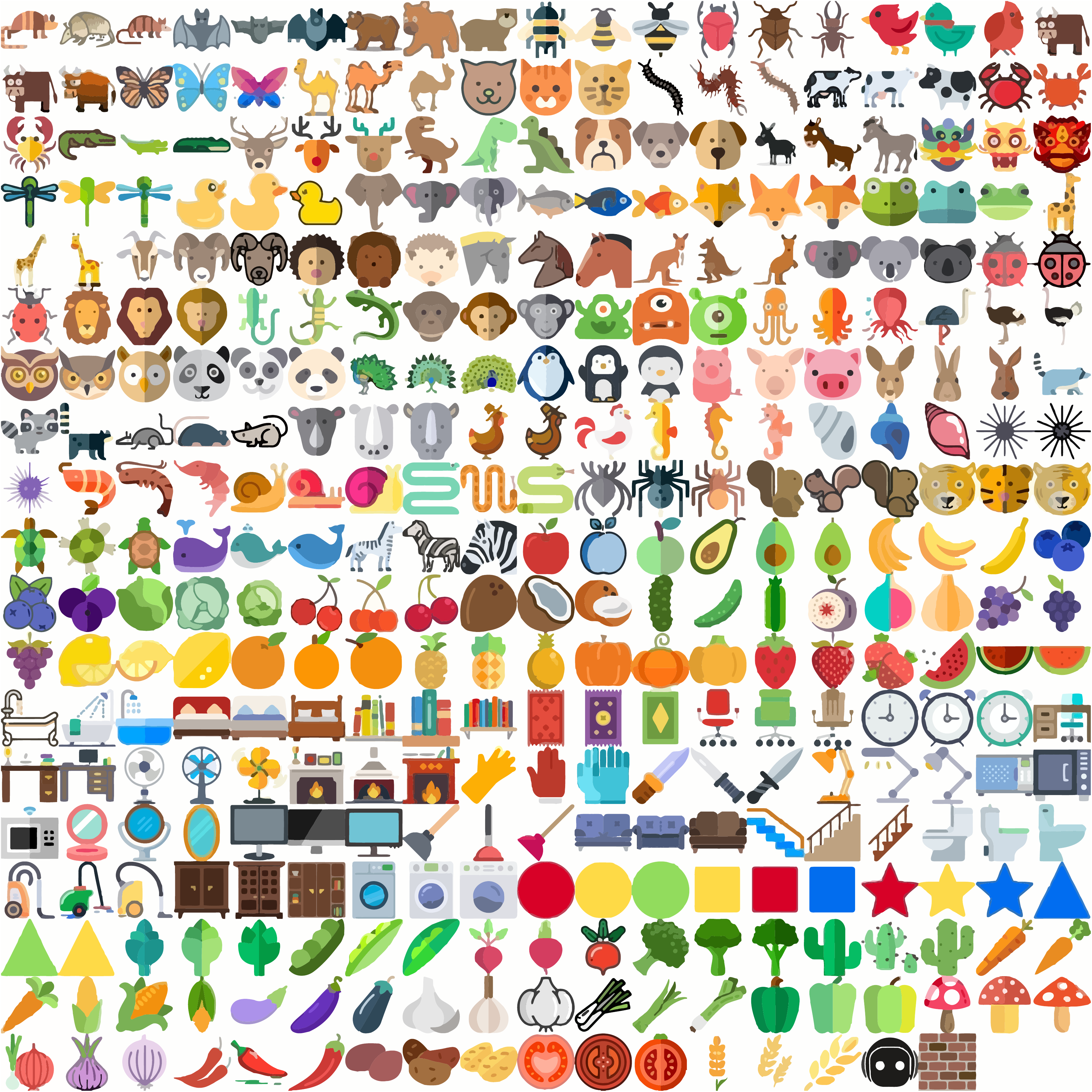}
    }
    \caption{All the $119\times 3=357$ object instances plus the agent
      (second-to-last) and the wall (last).
    }
    \label{fig:icons}
  \end{center}
\end{figure}

\section{\textsc{xworld} setup}
\label{app:xworld}
\textsc{xworld} is configured with $7\times 7$ grid
maps.
On each map, the open space for the agent has a size ranging from
$3\times 3$ to $7\times 7$.
Smaller open spaces are set up for curriculum learning
(Appendix~\ref{app:curriculum}), but not for testing.
To keep the size of the environment image fixed, we pad the map with
wall blocks if the open space has a size less than $7$.
The agent has four navigation actions in
total: \texttt{left}, \texttt{right}, \texttt{up}, and
\texttt{down}.
In each session,
\begin{enumerate}[I]
  \item The maximum number of time steps is four times the map size.
    That is, the agent only has $7\times 4=28$ steps to reach a target.
  \item The number of objects on the map ranges from 1 to 5.
  \item The number of wall blocks on the map ranges from 0 to 15.
  \item The positive reward when the agent reaches the correct
    location is $1.0$.
  \item The negative rewards for hitting walls and for stepping
    on non-target objects are $-0.2$ and $-1.0$, respectively.
  \item
    The time penalty of each step is $-0.1$.
\end{enumerate}

The teacher has a vocabulary size of 185.
There are 9 spatial relations, 8 colors, 119 distinct object
classes, and 50 grammatical words.
Every object class contains 3 different instances.
All object instances are shown in Figure~\ref{fig:icons}.
Every time the environment is reset, a number of object classes are
randomly sampled and an object instance is randomly sampled for each
class.
There are in total 16 types of sentences that the teacher can speak,
including 4 types of NAV commands and 12 types of QA questions.
Each sentence type has several non-recursive templates, and
corresponds to a concrete type of tasks the agent must learn to accomplish.
In total there are 1,639,015 distinct sentences with 567,579 for NAV and
1,071,436 for QA.
The sentence length varies between 2 and 13.

The object, spatial-relation, and color words of the teacher's
language are listed in Table~\ref{tab:words}.
These are the content words that can be grounded in \textsc{xworld}.
All the others are grammatical words.
Note that the differentiation between the content and the grammatical
words is automatically learned by the agent according to the tasks.
Every word is represented by an entry in the word embedding table.

\begin{table}[!t]
    \begin{center}
      \resizebox{\textwidth}{!}{
    \begin{tabular}{llll}
      \textbf{Object} & \textbf{Spatial relation} & \textbf{Color} & \textbf{Other}\\
      \hline\\
      apple, armadillo, artichoke, avocado, banana, bat,
      &
      between,
      &
      blue,
      &
      ?, ., and,
      \\
      bathtub, beans, bear, bed, bee, beet,
      &
      east,
      &
      brown,
      &
      block, by, can,
      \\
      beetle, bird, blueberry, bookshelf, broccoli, bull,
      &
      north,
      &
      gray,
      &
      color, could, destination,
      \\
      butterfly, cabbage, cactus, camel, carpet, carrot,
      &
      northeast,
      &
      green,
      &
      direction, does, find,
      \\
      cat, centipede, chair, cherry, circle, clock,
      &
      northwest,
      &
      orange,
      &
      go, goal, grid,
      \\
      coconut, corn, cow, crab, crocodile, cucumber,
      &
      south,
      &
      purple,
      &
      have, identify, in,
      \\
      deer, desk, dinosaur, dog, donkey, dragon,
      &
      southeast,
      &
      red,
      &
      is, locate, located,
      \\
      dragonfly, duck, eggplant, elephant, fan, fig,
      &
      southwest,
      &
      yellow
      &
      location, me, move,
      \\
      fireplace, fish, fox, frog, garlic, giraffe,
      &
      west
      &
      &
      name, navigate, near,
      \\
      glove, goat, grape, greenonion, greenpepper, hedgehog,
      &
      &
      &
      nothing, object, of,
      \\
      horse, kangaroo, knife, koala, ladybug, lemon,
      &
      &
      &
      on, one, please,
      \\
      light, lion, lizard, microwave, mirror, monitor,
      &
      &
      &
      property, reach, say,
      \\
      monkey, monster, mushroom, octopus, onion, orange,
      &
      &
      &
      side, target, tell,
      \\
      ostrich, owl, panda, peacock, penguin, pepper,
      &
      &
      &
      the, thing, three,
      \\
      pig, pineapple, plunger, potato, pumpkin, rabbit,
      &
      &
      &
      to, two, what,
      \\
      racoon, rat, rhinoceros, rooster, seahorse, seashell,
      &
      &
      &
      where, which, will,
      \\
      seaurchin, shrimp, snail, snake, sofa, spider,
      &
      &
      &
      you, your
      \\
      square, squirrel, stairs, star, strawberry, tiger,
      &
      &
      &
      \\
      toilet, tomato, triangle, turtle, vacuum, wardrobe,
      &
      &
      &
      \\
      washingmachine, watermelon, whale, wheat, zebra
      &
      &
      &
      \\
    \end{tabular}
  }
\end{center}
\caption{The teacher's lexicon.}
\label{tab:words}
\end{table}

The sentence types that the teacher can speak are listed in
Table~\ref{tab:sentence-types}.
Each type has a triggering condition about when the teacher says that
type of sentences.
Besides the shown conditions, an extra condition for NAV commands is
that the target must be reachable from the current agent location.
An extra condition for color-related questions is that the object
color must be one of the eight defined colors.
If at any time step there are multiple types triggered, we randomly
sample one for NAV and another for QA.
After a sentence type is sampled, we generate a sentence according to
the corresponding sentence templates.

\newcommand\Tstrut{\rule{0pt}{2.6ex}}         
\newcommand\Bstrut{\rule[-0.9ex]{0pt}{0pt}}   

\begin{table}[!t]
\begin{center}
\resizebox{\textwidth}{!}{
\begin{tabular}{lll}
  Sentence Type & Example & Triggering Condition \Bstrut\\
  \hline
  \texttt{nav\_obj} & ``Please go to the apple.''
  & [C0] Beginning of a session. \& \Tstrut\\
  & & [C1] The reference object has a unique\\
  & & name in the environment. \Bstrut\\
  \hline
  \texttt{nav\_col\_obj} & ``Could you please move to the
    red apple?'' &
  [C0] \& [C2] There are multiple objects \Tstrut\\
  & & that either have the same name but\\
  & & different colors, or have different\\
  & & names but the same color. \Bstrut\\
  \hline
  \texttt{nav\_nr\_obj} & ``The north of the apple is your
  destination.'' & [C0] \& [C1] \Tstrut\Bstrut\\
  \hline
  \texttt{nav\_bw\_obj} & ``Navigate to the grid
    between apple and &
  [C0] \& [C3] Both reference objects \Tstrut\\
  & banana please.'' & have unique names in the environment \\
  & & and are separated by one block. \Bstrut\\
  \hline
  \texttt{rec\_col2obj} & ``What is the red object?''
  & [C4] There is only one object that has \Tstrut\\
  & & the specified color. \Bstrut\\
  \hline
  \texttt{rec\_obj2col} & ``What is the color of the
  apple?'' & [C1] \Tstrut\Bstrut\\
  \hline
  \texttt{rec\_loc2obj} & ``Please tell the name of the
  object in the south.'' & [C5] The agent is near the reference \Tstrut\\
  & & object.\Bstrut\\
  \hline
  \texttt{rec\_obj2loc} & ``What is the location of the
  apple?'' & [C1] \& [C5] \Tstrut\Bstrut\\
  \hline
  \texttt{rec\_loc2col} & ``What color does the object in
  the east have?'' & [C5] \Tstrut\Bstrut\\
  \hline
  \texttt{rec\_col2loc} & ``Where is the red object located?''
  & [C4] \& [C5] \Tstrut\Bstrut\\
  \hline
  \texttt{rec\_loc\_obj2obj} & ``Identify the
    object which is in the east of the apple.''
  & [C1] \& [C6] There is an object near the \Tstrut\\
    & & reference object. \Bstrut\\
  \hline
  \texttt{rec\_loc\_obj2col} & ``What is the
  color of the east to the apple?'' & [C1] \& [C6] \Tstrut\Bstrut\\
  \hline
  \texttt{rec\_col\_obj2loc} & ``Where is the
  red apple?'' & [C2] \& [C5] \Tstrut\Bstrut\\
  \hline
  \texttt{rec\_bw\_obj2obj} & ``What is the
    object between apple and banana?'' &
  [C7] Both reference objects have unique \Tstrut\\
  & & names in the environment and are\\
  & & separated by a block. \Bstrut\\
  \hline
  \texttt{rec\_bw\_obj2loc} & ``Where is the
    object between apple and banana?''
  & [C7] \& [C8] The agent is near the block \Tstrut\\
    & & which is between the two reference\\
    & & objects. \Bstrut\\
  \hline
  \texttt{rec\_bw\_obj2col} & ``What is the
    color of the object between apple & [C7] \Tstrut\\
  & and banana?'' &\\
\end{tabular}
}
\end{center}
\caption{All the sixteen sentence types of the teacher.}
\label{tab:sentence-types}
\end{table}

\section{Implementation details}
\label{app:details}

The environment image~$e$ is a $156\times 156$ egocentric RGB
image.
The CNN in~$\mathbf{F}$ has four convolutional layers: $(3,3,64),
(2,2,64), (2,2,256), (1,1,256)$, where $(a,b,c)$ represents a layer
configuration of $c$ kernels of size $a$ applied at stride width $b$.
All the four layers are ReLU activated.
To enable the agent to reason about spatial-relation words (\eg\
``north''), we append an additional parametric cube to the
convolutional output to obtain~$h$.
This parametric cube has the same number of channels with the CNN
output, and it is initialized with a zero mean and a zero standard
deviation.

The word embedding table is initialized with a zero mean and a unit
standard deviation.
All the gated RNNs (including the bidirectional RNN) in~$\mathbf{L}$
have $128$ units.
All the layers in~$\mathbf{L}$, unless otherwise stated, use tanh as
the activation function.

For NAV, $x_{\text{loc}}$ is used as the target to reach on the image plane.
However, knowing this alone does not suffice.
The agent must also be aware of walls and possibly confounding targets
(other objects) in the environment.
Toward this end, $\mathbf{M}_A$ further computes an environment
terrain map $x_{\text{terr}}=\sigma(hf)$ where $f\in\mathbb{R}^D$ is a
parameter vector to be learned and~$\sigma$ is sigmoid.
We expect that~$x_{\text{terr}}$ detects any blocks informative for navigation.
Note that~$x_{\text{terr}}$ is unrelated to the specific command; it solely
depends on the current environment.
After stacking~$x_{\text{loc}}$ and~$x_{\text{terr}}$ together, $\mathbf{M}_A$ feeds
them to another CNN followed by an MLP.
The CNN has two convolutional layers $(3,1,64)$ and $(3,1,4)$, both
with paddings of $1$.
It is followed by a three-layer MLP where each layer has $512$ units
and is ReLU activated.

The action module~$\mathbf{A}$ contains a two-layer MLP of which the
first layer has $512$ ReLU activated units and the second layer is
softmax whose output dimension is equal to the number of actions.

We use adagrad \citep{Duchi2011} with a learning rate of $10^{-5}$ for
stochastic gradient descent (SGD).
The reward discount factor is set to $0.99$.
All the parameters have a default weight decay of~$10^{-4}\times 16$.
For each layer, its parameters have zero mean and a standard deviation
of $1 \mathbin{/} \sqrt{K}$, where $K$ is the number of parameters of
that layer.
We set the maximum interpretation step $I=3$.
The whole model is trained end to end.

\begin{figure}[t!]
  \begin{center}
    \includegraphics[width=0.7\textwidth]{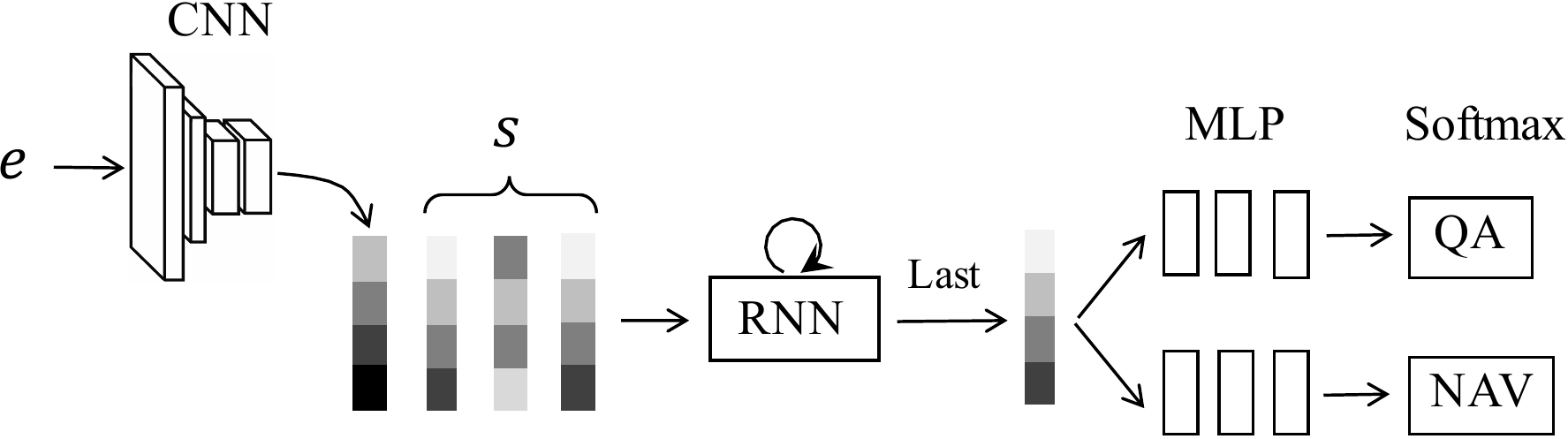}
    \caption{An overview of the baseline~\textbf{VL}.
      The computations of NAV and QA only differ in the last MLPs.
    }
    \label{fig:VL}
  \end{center}
\end{figure}

\begin{figure}[t!]
  \begin{center}
    \includegraphics[width=0.6\textwidth]{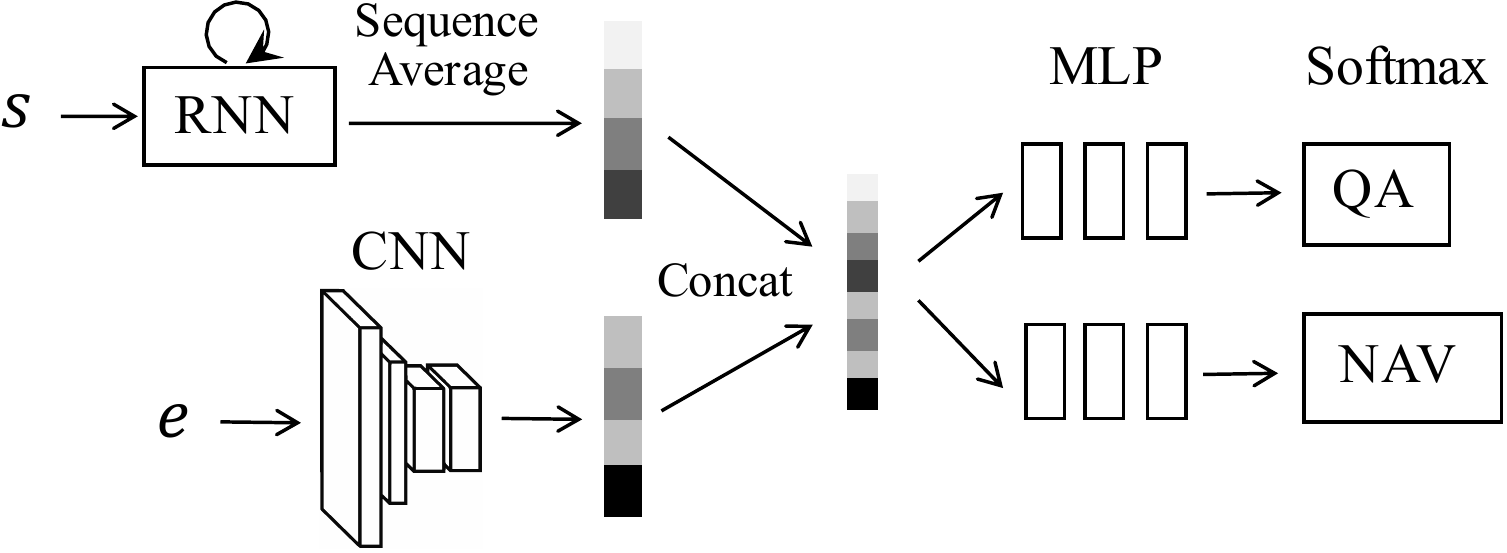}
    \caption{An overview of the baseline~\textbf{CE}.
      The computations of NAV and QA only differ in the last MLPs.
    }
    \label{fig:CE}
  \end{center}
\end{figure}

\section{Baseline details}
\label{app:baselines}
Some additional implementation details of the baselines in
Section~\ref{sec:baselines} are described below.
\begin{enumerate}[]
\item{\textbf{[CA]}} Its RNN has $512$ units.
\item{\textbf{[VL]}} Its CNN has four convolutional layers $(3,2,64)$,
  $(3,2,64)$, $(3,2,128)$, and $(3,1,128)$.
  This is followed by a fully-connected layer of size $512$, which
  projects the feature cube to the word embedding space.
  The RNN has $512$ units.
  For either QA or NAV, the RNN's last state goes through a three-layer
  MLP of which each layer has $512$ units (Figure~\ref{fig:VL}).
\item{\textbf{[CE]}} It has the same layer-size configuration with
  \textbf{VL} (Figure~\ref{fig:CE}).
\item{\textbf{[SAN]}} Its RNN has $256$ units.
  Following the original approach~\citep{Yang2016}, we use two
  attention layers.
\end{enumerate}

All the layers of the above baselines are ReLU activated.

\section{Exploration and Experience Replay}
\label{app:ac}
The agent has one million exploration steps in total, and the
exploration rate~$\lambda$ decreases linearly from $1$ to $0.1$.
At each time step, the agent takes an action~$a\in\{\text{\texttt{left}, \texttt{right},
\texttt{up}, \texttt{down}}\}$ with a probability of
\[\lambda\cdot\frac{1}{4} + (1-\lambda)\cdot \pi_{\theta}(a|s,e),\]
where~$\pi_{\theta}$ is the current policy, and $s$ and $e$ denote the
current command and environment image, respectively.
To stabilize the learning, we also employ experience replay
(ER) \citep{Mnih2015}.
The environment inputs, rewards, and the actions taken by the agent in
the most recent 10k time steps are stored in a replay buffer.
During training, each time a minibatch $\{a_i, s_i, e_i,
r_i\}_{i=1}^N$ is sampled from the buffer, using the rank-based
sampler \citep{Schaul2015} which has proven to increase the training
efficiency by prioritizing rare experiences.
Then we compute the gradient as:
\[-\sum_{i=0}^N\big(\nabla_{\theta}\log\pi_{\theta}(a_i|s_i,e_i)+\nabla_{\theta}v_{\theta}(s_i,e_i)\big)\big(r_i+\gamma
 v_{\theta^-}(s_i',e_i')-v_{\theta}(s_i,e_i)\big),\]
where $i$ is the sample index in the batch, $s_i'$ and $e_i'$ are the
command and image in the next time step, $v$ is the value function,
$\theta$ are the current parameters, $\theta^-$ are the target
parameters that have an update delay, and $\gamma$ is the discount
factor.
This gradient maximizes the expected reward while minimizing the
temporal-difference (TD) error.
Note that because of ER, our AC method is off-policy.
To avoid introducing biases into the gradient, importance ratios are
needed.
However, we ignored them in the above gradient for implementation
simplicity.
We found that the current implementation worked well in practice for
our problem.

\section{Curriculum learning}
\label{app:curriculum}
We exploit curriculum learning \citep{Bengio2009} to gradually
increase the environment complexity to help the agent learn.
The following quantities are increased in proportional to $\min(1, G'
\mathbin{/} G)$, where $G'$ is the number of sessions trained so far
and $G$ is the total number of curriculum sessions:
\begin{enumerate}[I]
\item The size of the open space on the environment map.
\item The number of objects in the environment.
\item The number of wall blocks.
\item The number of object classes that can be sampled from.
\item The lengths of the NAV command and the QA question.
\end{enumerate}
We found that this curriculum is important for an efficient learning.
Specifically, the gradual changes of quantities IV and V are supported
by the findings of \citet{Siskind1996} that children learn new words
in a linguistic corpus much faster after partial exposure to the corpus.
In the experiments, we set $G=$25k during training while do
\emph{not} have any curriculum during test (\ie\ testing with the maximum
difficulty).
